\documentclass[11pt]{article}

\usepackage[final]{acl}

\usepackage{times}
\usepackage{latexsym}

\usepackage[T1]{fontenc}

\usepackage{algorithm}
\usepackage{algpseudocode}
\usepackage{amsmath}
\usepackage{amssymb}
\usepackage{multirow}       
\usepackage{makecell}     
\usepackage{booktabs}     
\usepackage{paralist}
\usepackage{xcolor}
\usepackage{pifont}
\usepackage{footmisc}
\usepackage{pifont}

\usepackage{CJK}

\usepackage{graphicx}
\usepackage{multirow}
\usepackage{xcolor}  
\usepackage{color}
\usepackage{colortbl}
\usepackage[skins]{tcolorbox}
\usepackage{makecell}
\definecolor{lightblue}{RGB}{229,248,255}
\definecolor{gray}{rgb}{0.9,0.9,0.9}
\usepackage{microtype}

\usepackage{inconsolata}

\usepackage{graphicx}

\title{Ancient-Bench: A Comprehensive Multi-millennial, Multi-medium, and Multi-script Benchmark for Ancient Chinese Artifact Text Recognition}

\author{
 \textbf{Hiuyi Cheng\textsuperscript{1}},
 \textbf{Nuo Xu\textsuperscript{2}},
 \textbf{Yuyi Zhang\textsuperscript{1}},
 \textbf{Xuhan Zheng\textsuperscript{1}},
 \textbf{Wei Pan\textsuperscript{1}},
 \textbf{Jing Zhang\textsuperscript{2}},
\\
 \textbf{Dezhi Peng\textsuperscript{2}\thanks{Corresponding authors.}},
 \textbf{Minghui Liao\textsuperscript{2}},
 \textbf{Yihua Teng\textsuperscript{2}},
 \textbf{Jihao Wu\textsuperscript{2}},
 \textbf{Haoyu Ren\textsuperscript{2}},
 \textbf{Lianwen Jin\textsuperscript{1}\footnotemark[1]}
\\
 \textsuperscript{1}South China University of Technology,
 \textsuperscript{2}Huawei Technologies Co., Ltd.
\\
  \texttt{cheng.hiuyi329@gmail.com, pengdezhi5@huawei.com, eelwjin@scut.edu.cn} \\
  \texttt{\url{https://github.com/SCUT-DLVCLab/Ancient_Bench}}
 }
 
\begin{document}
\maketitle
\begin{abstract}
Ancient Chinese artifact text recognition is fundamental to heritage digitization, and benchmarks for ancient texts are essential for evaluating current model capabilities. However, existing benchmarks suffer from ``fragmentation'', manifested in limited temporal coverage, limited medium diversity, and incomplete script types. Therefore, we present \textbf{Ancient-Bench}, a comprehensive benchmark of 2,700 images for ancient Chinese artifact text recognition, featuring three dimensions: \textbf{Multi-millennial} (spanning 3,000 years of character evolution), \textbf{Multi-medium} (covering nine artifact categories), and \textbf{Multi-script} (encompassing seven historical script forms). To enable consistent and fair evaluation across heterogeneous media, we further define three annotation standards tailored to the medium-specific characteristics of ancient texts: symbol standardization, character standardization, and parsing standardization. 
Extensive experiments on \textbf{Ancient-Bench} covering general Vision-Language Models (VLMs) and OCR-specialist models reveal that ancient Chinese artifact text recognition remains fundamentally unsolved, with persistent challenges in variant characters, specialized symbols, and hallucination. 

\end{abstract}

\vspace{-2mm}
\section{Introduction}
\vspace{-2mm}

The digitization of ancient Chinese artifacts is fundamental to cultural heritage preservation and digital humanities research~\cite{intro}. Ancient texts inscribed on diverse physical media constitute the foundational documentary record of Chinese civilization spanning over three millennia. Accurate recognition of these artifacts is crucial for systematically studying the evolutionary trajectory of Chinese civilization and advancing the digital preservation of cultural heritage.

As shown in Table \ref{tab:comparison_benchmark}, existing benchmarks (\citealp{MTH1000, OBI-Bench, MCHDoc}) suffer from four limitations: \textbf{medium specificity} (most target a single artifact type in isolation), \textbf{script incompleteness} (none covers all Chinese script forms), \textbf{temporal sparsity} (limited time span/period coverage), and \textbf{absent annotation standards} (none rigorously defines conventions for medium-specific symbols, character variants, or structured layout parsing). This ``fragmentation problem'' hinders systematic evaluation of model generalization across eras, media, and scripts.

To address these challenges, we present \textbf{Ancient-Bench}, which contains 2,700 annotated images, spanning 3,000 years of character evolution, covering nine artifact categories from 14 national cultural institutions, and encompassing 7 historical script forms. In order to establish annotation across diverse artifact media, we define three annotation standards tailored to ancient artifacts: symbol standardization, character standardization, and parsing standardization.

Extensive experiments across general VLMs and OCR-specialist models reveal significant overall challenges, particularly on oracle bone and bronze inscriptions. Furthermore, in-depth analysis based on experimental results expose systematic deficiencies in variant character recognition and special symbol handling, demonstrating both the necessity and the difficulty of Ancient-Bench.

\begin{table*}[t]
\centering
\caption{Comparison of ancient artifact benchmarks across media types, script forms, and timeline coverage, illustrating the scope of Ancient-Bench.}
\vspace{-2mm}
\label{tab:comparison_benchmark}
\resizebox{0.93\linewidth}{!}{
    \begin{tabular}{lccccccccc}
    \hline
    \multirow{2}{*}{\textbf{Benchmark}} & \multicolumn{9}{c}{\textbf{Medium Types}} \\ \cline{2-10}
     & \textbf{Oracle} & \textbf{Bronze} & \textbf{Slip} & \textbf{Silk} & \textbf{Seal} & \textbf{Stele} & \textbf{Cliff} & \textbf{Edition} & \textbf{Calligraphy} \\ \hline
    MTH1000 (\citealp{MTH1000}) &  &  &  &  &  &  &  & \ding{51} &  \\
    M$^5$HisDoc (\citealp{M5HisDoc}) &  &  &  &  &  &  &  & \ding{51} &  \\
    HisDoc1B (\citealp{HisDoc1B}) &  &  &  &  &  &  &  & \ding{51} &  \\
    OBI-Bench (\citealp{OBI-Bench}) & \ding{51} &  &  &  &  &  &  &  &  \\
    DeepJianDu (\citealp{DeepJiandu}) &  &  & \ding{51} &  &  &  &  &  &  \\
    CalliBench (\citealp{CalliBench}) &  &  &  &  &  &  &  &  & \ding{51} \\
    MCHDoc (\citealp{MCHDoc}) & \ding{51} &  & \ding{51} & \ding{51} &  & \ding{51} &  & \ding{51} & \ding{51} \\ \hline
    \textbf{Ancient-Bench (Ours)} & \ding{51} & \ding{51} & \ding{51} & \ding{51} & \ding{51} & \ding{51} & \ding{51} & \ding{51} & \ding{51} \\ \hline
    \multirow{2}{*}{\textbf{Benchmark}} & \multicolumn{7}{c}{\textbf{Script Types}} & \multicolumn{2}{c}{\multirow{2}{*}{\textbf{Timeline}}} \\ \cline{2-8}
     & \textbf{Oracle Bone} & \textbf{Bronze} & \textbf{Seal} & \textbf{Clerical} & \textbf{Regular} & \textbf{Cursive} & \textbf{Running} & \multicolumn{2}{l}{} \\ \hline
    MTH1000 (\citealp{MTH1000}) &  &  &  &  & \ding{51} &  &  & \multicolumn{2}{l}{960-1911 CE} \\
    M$^5$HisDoc (\citealp{M5HisDoc}) &  &  &  & \ding{51} & \ding{51} & \ding{51} &  & \multicolumn{2}{l}{206 BCE-1911 CE} \\
    HisDoc1B (\citealp{HisDoc1B}) &  &  &  & \ding{51} & \ding{51} & \ding{51} & \ding{51} & \multicolumn{2}{l}{206 BCE-1911 CE} \\
    OBI-Bench (\citealp{OBI-Bench}) & \ding{51} &  &  &  &  &  &  & \multicolumn{2}{l}{1300-1046 BCE} \\
    DeepJianDu (\citealp{DeepJiandu}) &  &  & \ding{51} & \ding{51} &  &  &  & \multicolumn{2}{l}{221 BCE-220 CE} \\
    CalliBench (\citealp{CalliBench}) &  &  & \ding{51} & \ding{51} & \ding{51} & \ding{51} & \ding{51} & \multicolumn{2}{l}{221 BCE-1911 CE} \\
    MCHDoc (\citealp{MCHDoc}) & \ding{51} &  & \ding{51} & \ding{51} & \ding{51} & \ding{51} & \ding{51} & \multicolumn{2}{l}{1300 BCE-1911 CE} \\ \hline
    \textbf{Ancient-Bench (Ours)} & \ding{51} & \ding{51} & \ding{51} & \ding{51} & \ding{51} & \ding{51} & \ding{51} & \multicolumn{2}{l}{\textbf{1200 BCE-1911 CE}} \\ \hline
    \end{tabular}
}
\vspace{-4mm}
\end{table*}

In summary, the primary contributions of this paper are as follows:
\begin{itemize}
    \item We introduce \textbf{Ancient-Bench}, a comprehensive benchmark for ancient Chinese artifact text recognition that covers \textbf{9 media types, 7 script forms, and spans over 3,000 years}.
    \item We define three standardization protocols: symbol standardization, character standardization, and parsing standardization, tailored to the characteristics of ancient artifact media and knowledge systems.
    \item Extensive experiments on Ancient-Bench reveal that Ancient Chinese recognition remains fundamentally unsolved, with persistent challenges in handling variant characters and specialized symbols. 
\end{itemize}

\vspace{-2mm}
\section{Related work}
\vspace{-2mm}

\begin{figure*}[t]
  \centering
  \includegraphics[width=1\linewidth]{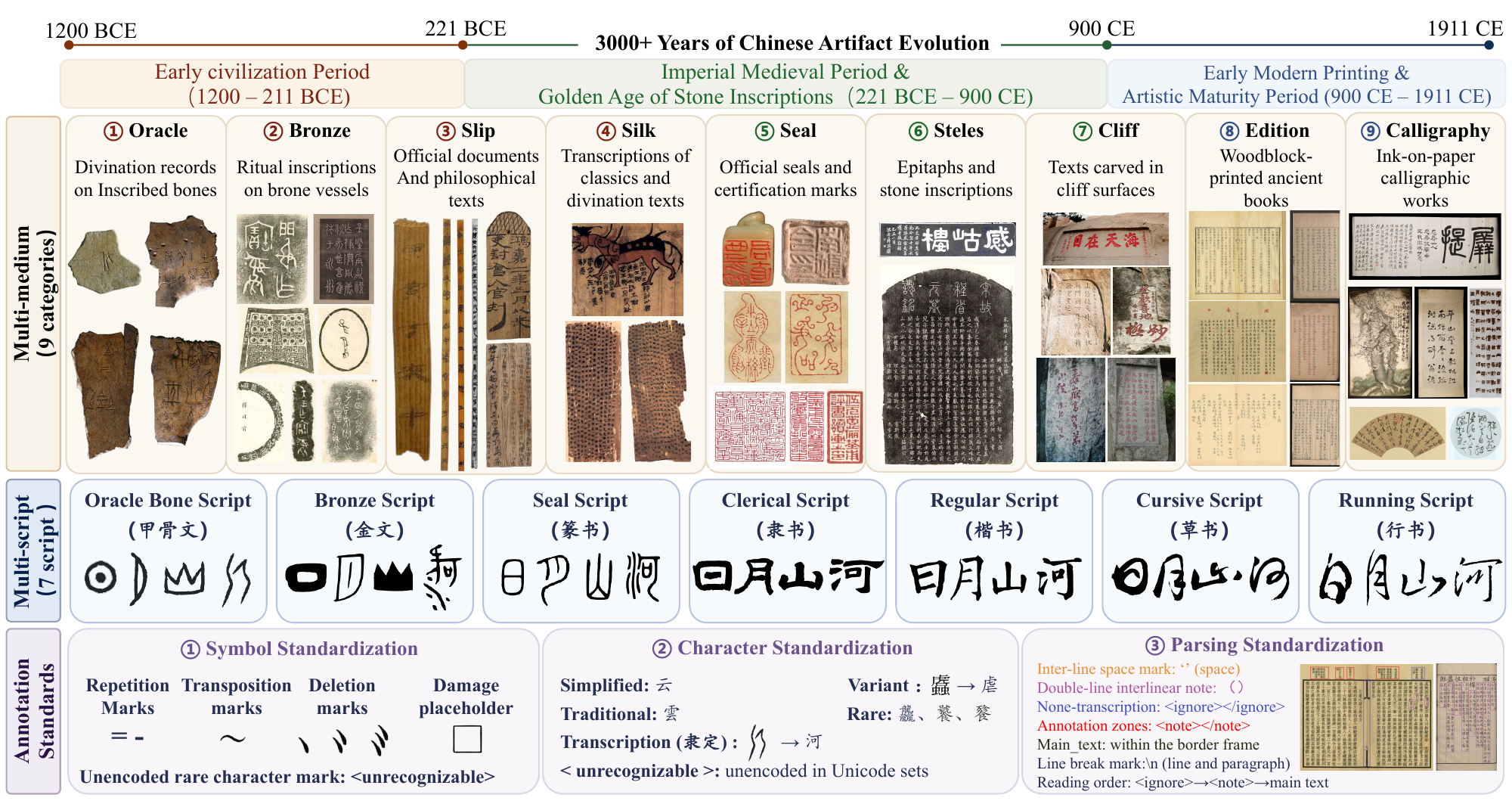}
  \vspace{-6mm}
  \caption{\textbf{Overview of the Ancient-Bench dataset.} Ancient-Bench comprises 2,700 meticulously annotated images, encompassing heterogeneous ancient artifact media, mainstream Chinese scripts, and over 3,000 years of character evolution. It provides character-level distinctions among paleographic transcriptions, simplified/traditional/rare characters, and variants, and additionally annotates special symbols, reading order, and layout information—establishing a unified standard for recognition across ancient Chinese artifact types.}
  \label{fig: main_figure}
  \vspace{-2mm}
\end{figure*}

In recent years, ancient artifact text recognition has attracted increasing attention, leading to the release of multiple specialized benchmark datasets. However, as shown in Table~\ref{tab:comparison_benchmark}, existing datasets remain limited in temporal span, medium diversity, script completeness, and systematic task definition.

\textbf{Single-medium datasets:} MTH1000 (\citealp{MTH1000}) was an early attempt targeting printed historical books, but it only covers a single script style (Regular script), with relatively homogeneous sample sources, and fails to encompass historical books from different periods and printing qualities. M$^5$HisDoc (\citealp{M5HisDoc}) expands to four script styles (Clerical, Regular, Cursive, and Running scripts) and introduces layout analysis and reading order tasks, covering more historical periods and printing versions. HisDoc1B (\citealp{HisDoc1B}) scales the data to billions of characters, with script coverage consistent with M$^5$HisDoc. Despite these advances, these datasets still suffer from severe medium limitations—they are entirely confined to paper-based printed historical books.

\textbf{Domain-specific datasets:} Some datasets focus on specific historical media or artistic forms. OBI-Bench (\citealp{OBI-Bench}) specializes in Shang Dynasty oracle bone script recognition, introducing layout parsing adapted to the morphology of turtle shells and animal bones, but it only covers oracle bone script. DeepJianDu (\citealp{DeepJiandu}) targets bamboo and wooden slips from the Qin and Han dynasties, capturing character form features during the seal--clerical script transition period, but is limited to slip materials from specific excavation sites, making it difficult to cover morphological variations of slips across different regions and periods. CalliBench (\citealp{CalliBench}) focuses on multi-script recognition in calligraphic artworks, covering various calligraphy styles but confined to the artistic creation domain. While these datasets delve deeply into a single medium each, they cannot support systematic evaluation across eras, media, and scripts.

\textbf{Multi-medium dataset attempts:} MCHDoc (\citealp{MCHDoc}) is a benchmark that attempts to encompass oracle bones, bamboo slips, silk manuscripts, stone inscriptions, historical books, and calligraphy. However, its data primarily originates from existing single-medium datasets, with inconsistent annotation standards; it lacks unified specifications for symbols specific to ancient artifacts (repetition marks, proofreading symbols, missing character marks, etc.) and fails to define cross-media differences in layout structure, reading order, and degradation characteristics.

Existing benchmarks generally suffer from a ``fragmentation'' problem: they operate independently across dimensions of medium, script, period, and task, lack unified annotation standards, and commonly overlook extensive structural and symbolic information in ancient artifacts. As a result, they cannot support ancient Chinese text recognition research across eras, media, and scripts.

\vspace{-3mm}
\section{Dataset Construction}
\vspace{-2mm}
\subsection{Design Principles}
\vspace{-2mm}

\begin{table}[!htbp]
\caption{Dataset Statistics for Each Medium Category}
\vspace{-2mm}
\label{tab: statistics}
\centering
\resizebox{0.95\linewidth}{!}{
\begin{tabular}{lcllc}
\hline
\textbf{Medium} & \textbf{Images} & \textbf{Resolution (px)} & \textbf{Chars}  & \textbf{Institution} \\ \hline
Oracle & 300 & 600$\times$800 & 1-68 & 1 \\
Bronze & 250 & 41$\times$80-1024$\times$1024 & 1-77 & 2 \\
Slip & 300 & 29$\times$224-700$\times$6820 & 1-107 & 3 \\
Silk & 300 & 46$\times$124-1330$\times$2067 & 1-784 & 2 \\ß
Seal & 350 & 100$\times$150-1333$\times$1341 & 1-40 & 4 \\
Steles & 300 & 85$\times$155-5795$\times$16745 & 4-816 & 4 \\
Cliff & 200 & 183$\times$113-7300$\times$5760 & 2-216 & Real World \\
Edition & 300 & 751$\times$506-970$\times$2817 & 4-776 & 1 \\
Calligraphy & 400 & 303$\times$509-4802$\times$13385 & 2-1666 & 4 \\ \hline
\textbf{Total} & 2700 & 41$\times$80-5795$\times$16745 & 1-1666 & 14 \\ \hline
\end{tabular}
}
\vspace{-5mm}
\end{table}

Existing research on ancient text recognition suffers from fragmented datasets, inconsistent standards, and insufficient spatiotemporal coverage. Most datasets focus on single media types or specific periods, hindering evaluation of model generalization in real-world scenarios. To address these challenges, we adopt a data-driven category formation strategy in constructing Ancient-Bench. We first extensively collect ancient images from authoritative cultural institutions (museums, libraries, etc.), and then derive the taxonomy of media, scripts, and periods from the artifacts' intrinsic characteristics, ensuring that Ancient-Bench authentically reflects the preservation status, distribution patterns, and recognition challenges of extant artifacts in real-world applications.

\vspace{-3mm}
\subsection{Data Sources and Collection}
\vspace{-1mm}

To ensure that Ancient-Bench is temporally comprehensive, diverse in media, and inclusive of script forms, we curate the dataset under strict provenance and quality requirements. To maintain the authority and broad coverage expected of an evaluation benchmark, we systematically reviewed authoritative sources and meticulously selected and integrated publicly accessible archival resources from 14 prestigious national-level cultural heritage institutions (e.g., museums and libraries), emphasizing cross-institutional diversity and representativeness of real-world preservation conditions.

During the data collection phase, we adhered to the following principles: (1) \textbf{Authoritative Sources:} All images were obtained from officially certified museums, libraries, and professional heritage institutions to ensure artifact authenticity and scholarly reliability; (2) \textbf{Diversity Assurance:} Within each institution, we broadly sampled artifacts across media, script forms, and historical periods, avoiding sampling bias toward specific categories; (3) \textbf{Quality Control:} We prioritized high-resolution, well-preserved images while retaining representative naturally degraded samples to reflect real-world recognition challenges.

\begin{figure*}[t]
\label{fig:stats}
  \includegraphics[width=0.95\linewidth]{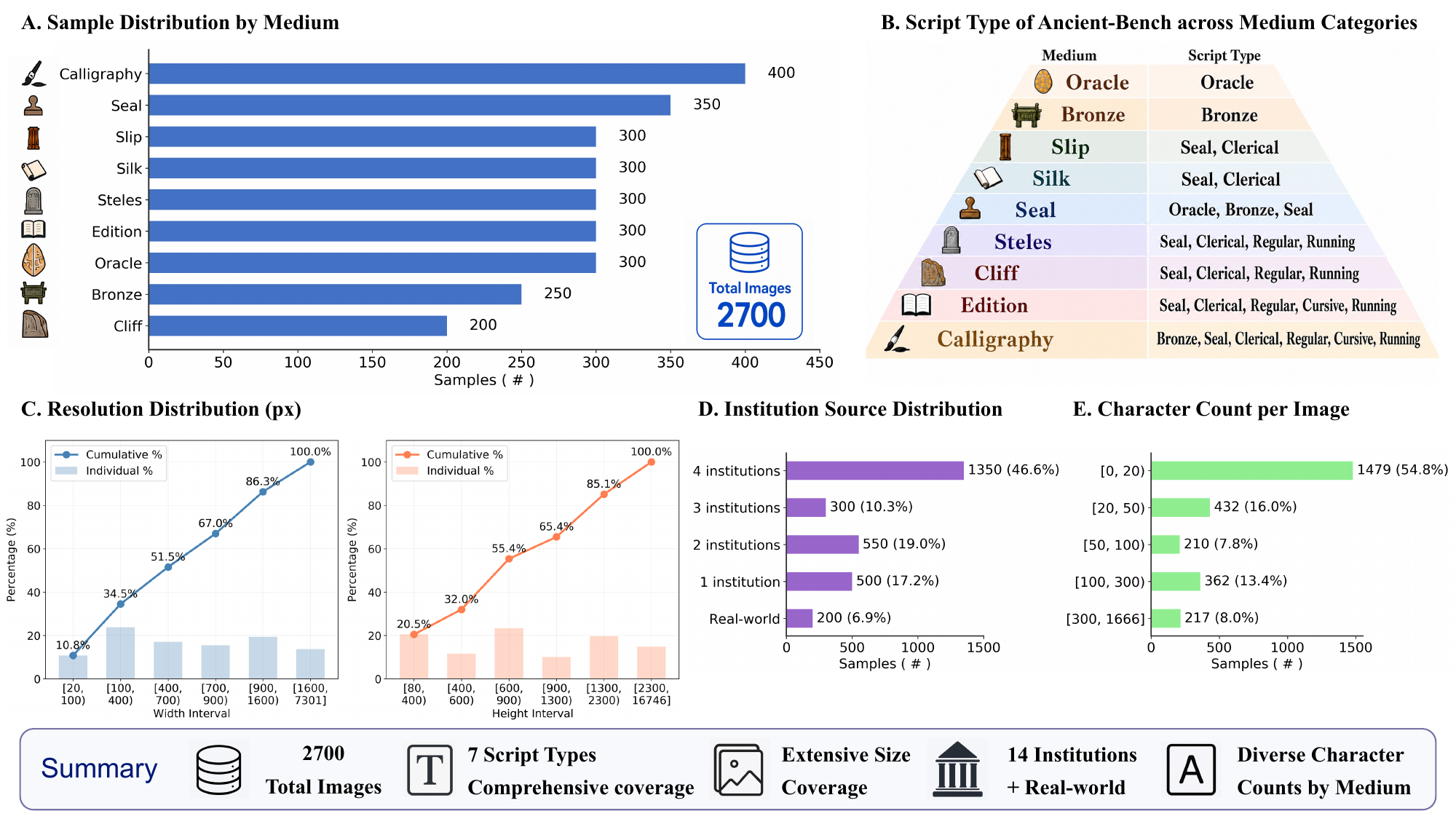}
  \centering
  \vspace{-2mm}
    \caption{\textbf{Overview of Ancient-Bench Dataset Statistics.}}
  \label{fig: dataset_statistics}
  \vspace{-2mm}
\end{figure*}

Through collection, processing, classification, and deduplication, we derived a taxonomy based on the actual distribution of historical resources. The resulting dataset encompasses \textbf{9 medium categories} (Oracle Bones (Oracle), Bronzes, Bamboo/Wooden Slips (Slip), Silk Manuscripts (Silk), Seals, Steles, Cliff Inscriptions (Cliff), Ancient Chinese Editions (Edition), and Calligraphy), \textbf{7 mainstream Chinese script forms} (Oracle Bone Script, Bronze Script, Seal Script, Clerical Script, Regular Script, Cursive Script, and Running Script), and \textbf{3 historical periods}. More detailed descriptions are provided in Appendix~\ref{sec:appendix_A1}.

\section{Ancient-Bench}
\vspace{-2mm}

This section provides an introduction to Ancient-Bench, with Figure~\ref{fig: main_figure} illustrating its data diversity.

\vspace{-2mm}
\subsection{Data Statistics}
\vspace{-1mm}

Ancient-Bench contains 2,700 meticulously annotated images spanning nine heterogeneous medium types, seven major Chinese script forms, and over 3,000 years of history. Figure~\ref{fig: dataset_statistics} provides a comprehensive statistical overview across five dimensions, while Table~\ref{tab: statistics} presents detailed statistics for each medium category, including character count ranges, image resolutions, and the number of source institutions.

\noindent \textbf{Regarding image resolution,} as shown in Figure~\ref{fig: dataset_statistics} (C), the cumulative distribution curves exhibit near-linear growth, indicating balanced coverage across resolution ranges. Resolutions span from 41$\times$80 pixels to 5{,}795$\times$16{,}745 pixels, covering two orders of magnitude and reflecting the physical characteristics of different media.

\noindent \textbf{Regarding text length,} as shown in Figure~\ref{fig: dataset_statistics} (E), the distribution is heavily skewed toward short texts. This directly reflects the physical constraints of ancient writing media: \textbf{short-text media} (seals, oracle bones, bronzes) typically range from 1--77 characters, constrained by limited surface area and engraving complexity; \textbf{medium-text media} (bamboo/wooden slips, cliff inscriptions, steles) range from 2--816 characters and are often used to record complete events or formal inscriptions; \textbf{long-text media} (silk manuscripts, editions, calligraphy) reach 361--1{,}666 characters, supporting full documentary transcription and artistic creation.

\noindent \textbf{Regarding data sources,} Ancient-Bench covers 14 national-level cultural heritage institutions, with 1--4 institutions per medium type. This multi-source design promotes (i) \textbf{geographical and stylistic diversity}, capturing variations across periods, regions, and scribes/artisans, and (ii) \textbf{preservation variance}, ranging from intact artifacts to heavily damaged samples, thereby reflecting real-world preservation conditions.

\noindent \textbf{Regarding script categories,} Ancient-Bench encompasses the complete evolutionary trajectory of Chinese script forms.

\vspace{-3mm}
\subsection{Data Processing Pipeline}

The Ancient-Bench dataset is sourced from major national-level cultural heritage institutions, whose original data and annotations vary substantially in standards, formats, and quality, making them unsuitable for rigorous and fair ancient text OCR evaluation. The core issues include:

\noindent \textbf{Inconsistent data formats.} Images vary in structure: some contain only text regions, while others include annotations, multiple regions (e.g., bamboo slips), or additional elements such as watermarks, artifact numbers, and background clutter, all of which can interfere with recognition.

\noindent \textbf{Annotations prioritize readability over fidelity.} Museum annotations are often simplified for modern readers rather than faithfully reflecting original appearances: (1) extensive use of simplified characters; (2) mixed simplified/traditional forms; (3) missing or damaged characters left unmarked; (4) variant characters normalized to simplified forms.

\noindent \textbf{Inconsistent annotation symbols.} Institutions adopt varying conventions to denote damage, repeated characters, inversions, deletions, and illegible content, leading to ambiguity and hindering reproducibility.

\vspace{-2mm}
\subsubsection{Image Pre-processing}
To address inconsistent data formats, we apply a three-step processing pipeline: (1) \textbf{Format standardization:} We batch-convert original PDF files into high-resolution images; (2) \textbf{MLLM-based quality filtering:} We leverage MLLMs to identify and filter images with prominent watermarks or irreversible physical damage. Since degradation is an inherent characteristic of silk manuscripts due to the medium itself, this category is exempt from damage-based filtering; and (3) \textbf{Precise region localization:} We use digital image processing algorithms to localize and crop regions of interest (ROIs) for text recognition, removing edge interference such as watermarks and artifact numbering.

\vspace{-2mm}
\subsubsection{Symbol Standardization}

Ancient Chinese text contain symbols with specific meanings that carry rich information and are of high scholarly value for textual criticism. Accordingly, we establish a cross-medium, standardized annotation standard that strictly adheres to the ``what you see is what you get'' principle.

\noindent \textbf{Repetition Marks (=, -).} In bamboo/wooden slips and silk manuscripts, ``='' and ``-'' can carry multiple meanings (e.g., repetition, ligature, proper names, or abbreviation). We therefore preserve them as-is to support scholarly verification. For consistency, repetition marks in other media are also retained unchanged.

\noindent \textbf{Damage Placeholder ($\square$).} This symbol is used when individual characters are blurred or damaged beyond recognition. For silk manuscripts with extensive continuous damage, where the number of missing characters cannot be determined, a single $\square$ is used to denote the entire damaged region.

\noindent \textbf{Unencoded Rare Character mark (<unrecognizable>}). This tag is applied uniformly to ancient forms, variants, and rare characters that are not encoded in Unicode.

\vspace{-2mm}
\subsubsection{Character Standardization}

Chinese characters exist in multiple forms, including clerical transcriptions of oracle bone and bronze inscriptions, simplified and traditional characters, as well as variant and rare characters. To ensure annotation accuracy and consistency, we establish a set of character standardization rules. The core objective of standardization is to maximize the scholarly value of ancient texts. Faithfully recording original character forms helps avoid semantic confusion and misinterpretation, enables researchers to examine historical usage patterns, and preserves temporal characteristics and evidence of textual evolution in artifacts, thereby providing a reliable data foundation for philology, textual criticism, and digital humanities research.

\noindent \textbf{Transcription of Ancient Scripts.} Characters in oracle bone and bronze inscriptions are archaic and often abstruse in form, differing substantially from modern Chinese characters and thus difficult to recognize. The standard practice in the field is \emph{transcription} (\begin{CJK}{UTF8}{gkai}隶定\end{CJK}), which converts ancient character forms into clerical or regular-script forms that are readable to modern readers. Such transcription should follow recognized academic standards, using traditional characters as the reference script to maintain scholarly consensus on form--meaning correspondences. Arbitrary replacement with simplified characters would disrupt the established correspondence system and deviate from accepted norms for interpreting ancient scripts. For example, the pictographic character for ``horse'' (\begin{CJK}{UTF8}{gkai}马\end{CJK}) inscribed on Shang-dynasty oracle bones is annotated using the transcribed character \begin{CJK}{UTF8}{bsmi}馬\end{CJK}.

\noindent \textbf{Principles for Handling Simplified and Traditional Characters.} Many Chinese characters were merged during character simplification; however, in Classical Chinese, the original forms often carried distinct meanings and usage patterns. For example, \begin{CJK}{UTF8}{bsmi}後\end{CJK} (``behind/after'') and \begin{CJK}{UTF8}{gkai}后\end{CJK} (``empress'') were merged into \begin{CJK}{UTF8}{gkai}后\end{CJK} in simplified Chinese, yet their meanings in ancient texts are entirely distinct. Therefore, we adopt the ``what you see is what you get'' principle and annotate according to the actual character forms in the image (simplified or traditional). This rule helps avoid semantic bias in interpretation.

\noindent \textbf{Principles for Handling Archaic Characters.} Slips, silk manuscripts, and steles often contain archaic characters, regional vernacular characters, and idiosyncratic forms that are not yet encoded in Unicode. We adopt two processing strategies: (1) \textbf{Variant Replacement:} If an unencoded character has an attested variant form and the institution's annotation permits replacement with that variant, we apply the replacement principle. For example, a complex archaic character with the structure \begin{CJK}{UTF8}{gkai}虎口虫虫\end{CJK} in the image does not exist in the character set, but its variant is \begin{CJK}{UTF8}{gkai}虐\end{CJK} (a standard regular-script character), so it is annotated as \begin{CJK}{UTF8}{gkai}虐\end{CJK}. This procedure requires confirming semantic consistency between the archaic character and its variant; (2) \textbf{Placeholder Replacement:} If the unencoded character cannot be entered and no corresponding variant or common character can be identified (e.g., hapax legomena from excavated slips or a calligrapher's unique scribal variants), we use the \texttt{<unrecognizable>} tag as a placeholder. This marker indicates that a character exists at the corresponding position in the original text but cannot be typed due to character set limitations.

\noindent \textbf{Principles for Handling Rare Characters.} Ancient texts contain rare characters, including historical rarities, name characters, obscure variants, and documentary \emph{hapax legomena} (character forms unique to artifacts). Many of these are encoded in Unicode extension blocks. We follow the WYSIWYG principle in annotation. For example: \begin{CJK}{UTF8}{bsmi}龘\end{CJK}.

\vspace{-2mm}
\subsubsection{Parsing Standardization}

The layout structure of ancient texts carries essential organizational logic and reading cues. Layout features such as line breaks in vertical writing, the semantic function of inter-line spacing, and the hierarchical relationships among marginal annotations are critical for understanding document content. To enable standardized parsing and annotation, we establish layout parsing standards that restore the original text arrangement and compositional logic, thereby preserving the typographic features and textual hierarchy of ancient texts.

\noindent \textbf{Line Break Mark (\texttt{\textbackslash n}).} We strictly restore line structure according to the original layout and reading order, uniformly marking natural line breaks in vertical text with \texttt{\textbackslash n}.

\noindent \textbf{Inter-line Spacing Mark (<space>)}. Spacing between columns in vertical text distinguishes columns and helps organize the text. Short spacing indicates phrasal pauses, whereas extensive spacing separates chapters, paragraphs, and texts, functioning similarly to modern paragraph breaks.

\noindent \textbf{Double-line Interlinear Note Symbol (\begin{CJK}{UTF8}{gkai}$\text{（）}$\end{CJK}).} Ancient texts commonly contain double-line, small-character interlinear notes, where smaller annotation text is inserted between lines of the main text to provide supplementary explanations, citation sources, textual variants, etc. We use parentheses \begin{CJK}{UTF8}{gkai}$\text{（）}$\end{CJK} to mark interlinear note content.

\noindent \textbf{Reading Order Rule.} Vertical text is read from right to left and from top to bottom; horizontal text is read from right to left. For cliff inscriptions, the reading direction follows that of body text (i.e., the portion with the length and largest character size).

\noindent \textbf{Printed Edition Layout Element Annotation Rule.} Editions contain non-transcription zones and annotation-extraction zones beyond the main text area (body text and double-line interlinear content), which would interfere with normal reading if left unprocessed. We therefore apply the following procedure: (1) \textbf{Non-transcription zones} (center seam, book ears) are enclosed in \texttt{<ignore></ignore>} tags in the order ``center seam $\rightarrow$ book ears,'' with line breaks by column; (2) \textbf{Annotation zones} (top margin, bottom margin, side notes, marginal notes) are enclosed in \texttt{<note></note>} tags in counterclockwise order, ``top $\rightarrow$ left $\rightarrow$ bottom $\rightarrow$ right,'' with line breaks by column.

\noindent \textbf{Seal Omission Principle}. Except for Seal dataset, collector and connoisseur seals from later periods are not annotated across all other media types.

\vspace{-3mm}
\subsection{Dataset Annotation}

\begin{table*}[t]
\centering
\vspace{-2mm}
\caption{\textbf{Evaluation results on ancient artifact recognition task.} \textbf{Bold} denotes SOTA (State-of-the-Art) performance, and $\underline{underline}$ indicates second-best performance. Results are reported separately for General VLMs and OCR Specialist Models.}
\vspace{-2mm}
\label{tab: evaluation}
\resizebox{\textwidth}{!}{
\begin{tabular}{lcccccccccc}
\hline
\begin{tabular}[c]{@{}l@{}}\textbf{Model}\\  (NED(\%) $\uparrow $ / F1(\%) $\uparrow$ )\end{tabular} & \textbf{Oracle} & \textbf{Bronze} & \textbf{Slip} & \textbf{Silk} & \textbf{Seal} & \textbf{Stele} & \textbf{Cliff} & \textbf{Editions} & \textbf{Calligraphy} & \textbf{Overall}  \\ \hline
\rowcolor{gray}
\multicolumn{11}{c}{\textit{\textbf{General VLMs: Closed-Source Models}}} \\ 
GPT-5-2025-08-07 \cite{GPT_5_2025} & 0.79 / 1.33 & 3.59 / 4.81 & 7.23 / 8.70 & 15.27 / 19.39 & 4.25 / 5.69 & 26.92 / 35.77 & 36.36 / 46.14 & 21.42 / 29.60 & 28.15 / 35.79 & 16.00 / 20.80 \\
GPT-4o-2024-08-06 \cite{GPT_4o_2023} & 0.89 / 2.05 & 5.60 / 10.39 & 11.37 / 14.22 & 15.84 / 22.50 & 3.60 / 6.46 & 23.72 / 43.03 & 30.41 / 45.80 & 22.98 / 44.42 & 28.63 / 39.10 & 15.90 / 25.33 \\
Gemini-3.1-pro-preview \cite{Gemini_3_1_2026} & \textbf{10.31} / \textbf{15.75} & 26.47 / 31.16 & 34.03 / 36.83 & 39.60 / 50.28 & 16.96 / 19.20 & 81.24 / 86.12 & 64.48 / 74.96 & 90.88 / 93.13 & 67.57 / 73.40 & 47.95 / 53.42 \\
Gemini-3.5-flash \cite{Gemini-3.5-flash} & 6.20 / 10.98 & 17.29 / 21.96 & 29.41 / 32.11 & 33.18 / 41.62 & 13.69 / 16.09 & 75.96 / 82.80 & 61.90 / 71.87 & 89.11 / 92.07 & 61.91 / 67.90 & 43.18 / 48.60 \\
Claude-Opus-4-7 \cite{Claude_Opus_4_7_2026} & 4.73 / 8.62 & 10.74 / 15.79 & 23.55 / 26.59 & 28.65 / 40.57 & 8.38 / 11.01 & 51.10 / 69.37 & 51.89 / 64.22 & 78.34 / 88.32 & 58.74 / 65.35 & 35.12 / 43.31 \\
Qwen3.6-Plus-2026-04-02 \cite{qwen3.6plus}  & 5.13 / 9.64 & 15.51 / 21.20 & 32.64 / 34.93 & 32.17 / 38.66 & 25.03 / 28.82 & 76.28 / 81.80 & 62.89 / 73.51 & 88.13 / 91.22 & 69.58 / 74.71 & 45.26 / 50.50 \\
Qwen3.6-Flash-2026-04-16 \cite{qwen3.5} & 4.42 / 8.11 & 15.33 / 23.20 & 26.95 / 31.55 & 29.27 / 38.59 & 18.43 / 22.67 & 72.90 / 80.02 & 60.78 / 71.84 & 82.75 / 88.47 & 67.44 / 73.77 & 42.03 / 48.69 \\
Doubao-seed-2-0-pro-260215 \cite{Seed_2_0_2026} & 3.82 / 6.22 & \underline{28.71} / 32.51 & 31.89 / 34.67 & 43.07 / 50.14 & \underline{34.61} / \underline{38.96} & \textbf{82.21} / \underline{88.54} & \textbf{68.76} / \textbf{78.27} & \textbf{93.14} / \textbf{94.66} & \underline{77.44} / \underline{83.63} & \underline{51.52} / 56.40 \\
Doubao-seed-2-0-lite-260428 \cite{Seed_2_0_2026} & 6.15 / 11.18 & \textbf{34.03} / \textbf{38.48} & 35.06 / 37.64 & \underline{44.34} / \textbf{52.21} & 32.11 / 38.60 & \underline{82.01} / \textbf{88.88} & 65.83 / \underline{76.92} & \underline{92.71} / \underline{94.49} & 76.51 / 82.39 & \textbf{52.08} / \textbf{57.86} \\
GLM-5V-Turbo \cite{GLM-5V-Turbo} & 3.42 / 6.81 & 7.66 / 11.99 & 22.89 / 25.22 & 29.67 / 36.86 & 10.30 / 13.77 & 57.63 / 65.93 & 47.47 / 57.44 & 81.74 / 86.60 & 53.53 / 58.87 & 34.92 / 40.39 \\  \hline
\rowcolor{gray}
\multicolumn{11}{c}{\textit{\textbf{General VLMs: Open-Source Models}}} \\ 
Qwen3.6-35B-A3B \cite{qwen3.6-35b-a3b} & 1.80 / 3.48 & 15.23 / 20.19 & 35.53 / 38.97 & 37.86 / 45.85 & 20.23 / 25.56 & 77.85 / 84.86 & \underline{66.54} / 76.54 & 83.03 / 90.19 & 75.61 / 81.47 & 45.96 / 51.90 \\
Qwen3.6-27B \cite{qwen3.6-27b} & 4.05 / 7.35 & 19.59 / 24.78 & \textbf{36.53} / 38.95 & \textbf{44.61} / \underline{50.89} & 22.43 / 27.71 & 77.39 / 83.95 & 65.35 / 74.28 & 87.03 / 91.29 & 74.93 / 80.11 & 47.99 / 53.26 \\
Qwen3.5-397B \cite{qwen3.5} & 3.94 / 7.06 & 14.86 / 19.85 & \underline{36.07} / \textbf{39.14} & 32.93 / 38.96 & 26.43 / 30.76 & 78.75 / 85.32 & 62.68 / 75.27 & 87.24 / 91.91 & 76.17 / 80.49 & 46.56 / 52.08 \\
Qwen3.5-2B \cite{qwen3.5} & 1.17 / 2.41 & 6.57 / 9.24 & 22.47 / 24.75 & 21.00 / 24.97 & 12.37 / 16.44 & 37.58 / 45.33 & 47.71 / 57.91 & 43.02 / 51.23 & 52.45 / 58.76 & 27.15 / 32.34 \\
InternVL3.5-241B-A28B \cite{internvl3.5} & 3.94 / 7.87 & 14.16 / 23.78 & 24.22 / 27.72 & 34.86 / 42.90 & 14.42 / 23.58 & 71.29 / 81.68 & 44.55 / 67.54 & 85.34 / 91.25 & 60.55 / 68.57 & 39.26 / 48.32 \\
InternVL3.5-2B \cite{internvl3.5} & 0.83 / 1.34 & 5.58 / 9.80 & 6.67 / 9.48 & 16.09 / 22.85 & 5.02 / 12.32 & 25.28 / 42.74 & 0.61 / 2.13 & 47.51 / 72.72 & 29.02 / 49.97 & 15.18 / 24.82 \\ 
GLM-4.6V-106B \cite{GLM-4.6V} & 1.14 / 2.05 & 4.77 / 7.57 & 17.54 / 21.91 & 20.73 / 25.46 & 4.87 / 9.87 & 45.30 / 64.27 & 45.56 / 58.55 & 68.48 / 80.43 & 51.66 / 56.46 & 28.89 / 36.29 \\
GLM-4.6V-Flash-9B \cite{GLM-4.6V} & 1.65 / 3.14 & 4.36 / 6.38 & 19.35 / 21.38 & 28.06 / 35.72 & 5.01 / 8.28 & 41.39 / 61.46 & 35.49 / 54.42 & 67.34 / 81.04 & 48.75 / 55.51 & 27.93 / 36.37 \\
Gemma-4-E2B-it \cite{Gemma_4} & 0.95 / 2.18 & 3.24 / 5.74 & 4.84 / 7.47 & 6.00 / 10.29 & 1.03 / 2.36 & 4.44 / 24.38 & 11.67 / 23.30 & 9.31 / 18.84 & 6.15 / 15.50 & 5.29 / 12.23 \\
Gemma-4-E4B-it \cite{Gemma_4} & 1.10 / 2.42 & 0.74 / 1.54 & 6.85 / 9.37 & 7.54 / 11.62 & 1.62 / 2.73 & 5.90 / 26.51 & 14.23 / 28.55 & 25.11 / 35.92 & 11.09 / 22.86 & 8.24 / 15.72 \\
Gemma-4-31B-it \cite{Gemma_4} & 0.35 / 0.48 & 2.49 / 3.12 & 16.80 / 18.79 & 17.73 / 19.22 & 4.63 / 5.99 & 41.27 / 51.04 & 43.01 / 51.71 & 45.39 / 53.11 & 38.93 / 44.62 & 23.40 / 27.56 \\
MiniCPM-V 4.6 \cite{minicpmv4.5} & 0.14 / 0.49 & 1.37 / 2.63 & 1.98 / 3.57 & 7.71 / 12.73 & 1.70 / 3.19 & 37.14 / 48.83 & 0.49 / 2.00 & 45.66 / 57.22 & 29.67 / 36.87 & 13.99 / 18.62 \\
Kimi-K2.6 \cite{kimi-k2.6} & \underline{8.66} / \underline{13.14} & 27.51 / \underline{34.73} & 35.27 / \underline{39.04} & 35.84 / 45.53 & \textbf{37.93} / \textbf{44.56} & 75.10 / 82.28 & 64.96 / 72.75 & 90.28 / 93.30 & \textbf{80.05} / \textbf{84.24} & 50.62 / \underline{56.62} \\ \hline
\rowcolor{gray}
\multicolumn{11}{c}{\textit{\textbf{OCR-Specialist VLMs: End-to-End Models}}} \\ 
Deepseek-OCR2 \cite{deepseek-ocr2} & 0.14 / 0.27 & 0.61 / 1.18 & 3.73 / 5.51 & 4.85 / 7.10 & 0.64 / 1.09 & 9.25 / 18.90 & 10.57 / 18.72 & 20.10 / 31.93 & 16.88 / 24.98 & 7.42 / 12.19 \\
HunyuanOCR \cite{hunyuanocr} & \textbf{1.42} / \underline{2.29} & \textbf{6.67} / \textbf{9.03} & \textbf{30.14} / \textbf{32.23} & \textbf{31.53} / \textbf{37.28} & \textbf{12.82} / \textbf{18.59} & \textbf{70.30} / \textbf{81.53} & \textbf{61.82} / \textbf{73.65} & \textbf{86.50} / \textbf{90.70} & \textbf{62.57} / \textbf{69.66} & \textbf{40.42} / \textbf{46.11} \\
OCRVerse \cite{OCRVerse} & 0.03 / 0.03 & 2.62 / 3.89 & 23.24 / \underline{25.75} & 21.01 / 26.03 & 6.20 / 8.41 & 49.82 / 62.58 & 37.28 / 53.36 & 63.54 / 79.99 & 45.92 / 55.92 & 27.74 / 35.11 \\
Qianfan-OCR \cite{qianfanocr} & 1.15 / 2.22 & 3.48 / 5.18 & 8.09 / 10.23 & 16.21 / 24.72 & 4.80 / 7.34 & 62.84 / \underline{72.51} & 32.60 / 51.13 & 81.24 / 88.10 & 46.94 / 54.53 & 28.59 / 35.11 \\
dots.ocr \cite{dotsocr} & 0.13 / 0.27 & 2.52 / 3.73 & 19.00 / 21.19 & 19.11 / 22.56 & 3.81 / 5.04 & 53.99 / 62.36 & \underline{43.00} / 55.52 & 84.30 / 89.82 & 50.91 / 58.41 & 30.75 / 35.43 \\
dots.mocr \cite{dotsmocr} & 0.00 / 0.00 & 2.64 / 3.96 & 18.60 / 20.71 & 20.06 / 23.80 & 7.55 / 9.23 & 46.11 / 62.87 & 31.07 / 43.02 & 63.53 / 80.11 & 39.09 / 51.01 & 25.41 / 32.75 \\
FireRed-OCR \cite{fireredocr} & 0.25 / 0.40 & 3.42 / 4.62 & 18.60 / 20.89 & 10.41 / 13.47 & 6.84 / 9.26 & 39.77 / 50.55 & 20.69 / 27.24 & 78.14 / 84.21 & 50.36 / \underline{58.75} & 25.39 / 29.93 \\
Nanonets-OCR2-3B \cite{Nanonets-OCR2} & 0.04 / 0.09 & 3.02 / 4.10 & 17.62 / 19.83 & 16.58 / 18.67 & 2.43 / 4.33 & 12.49 / 18.25 & 27.57 / 44.17 & 60.85 / 71.61 & 32.81 / 39.86 & 19.27 / 24.55 \\ \hline
\rowcolor{gray}
\multicolumn{11}{c}{\textit{\textbf{OCR-Specialist VLMs: Pipeline Models}}} \\ 
PaddleOCR-VL1.5 \cite{paddleocr-vl-1.5} & 1.04 / 1.71 & 3.48 / 4.56 & 20.45 / 22.41 & 26.08 / 32.03 & 4.00 / 6.41 & 61.68 / 70.89 & 32.19 / 52.37 & \underline{84.86} / \underline{90.64} & 48.84 / 56.36 & 31.40 / 37.49 \\
MinerU 2.5 Pro \cite{mineru25pro}  & 0.98 / 1.35 & 3.49 / 4.82 & 22.46 / 24.77 & 29.65 / \underline{36.01} & 4.92 / 6.83 & 57.51 / 66.64 & 33.85 / 53.71 & 80.41 / 87.25 & 45.30 / 52.72 & 30.95 / 37.12 \\
GLM-OCR \cite{glm-ocr} & \underline{1.40} / \textbf{2.45} & \underline{4.72} / \underline{5.84} & \underline{23.73} / 25.52 & \underline{31.03} / 35.24 & \underline{11.25} / \underline{14.34} & 61.43 / 68.74 & 42.10 / \underline{62.59} & 80.53 / 86.48 & \underline{53.59} / 58.61 & \underline{34.42} / \underline{39.98} \\
MonkeyOCR-3B \cite{monkeyocr} & 0.44 / 0.82 & 2.40 / 3.73 & 8.76 / 10.63 & 5.92 / 8.40 & 3.21 / 5.18 & 42.28 / 49.60 & 26.79 / 39.24 & 57.52 / 63.63 & 34.06 / 39.90 & 20.15 / 24.57 \\
MonkeyOCR-pro-3B \cite{monkeyocr} & 0.76 / 1.25 & 2.30 / 3.56 & 9.15 / 11.28 & 6.85 / 8.94 & 3.01 / 4.81 & 39.03 / 46.21 & 25.83 / 37.61 & 57.12 / 63.83 & 32.31 / 37.52 & 19.60 / 23.89 \\
YouTu-Parsinig \cite{youtuparsing} & 0.72 / 1.37 & 2.10 / 3.07 & 14.37 / 17.38 & 20.33 / 26.77 & 1.95 / 3.93 & 22.15 / 39.13 & 27.79 / 46.63 & 70.08 / 77.92 & 35.69 / 43.89 & 21.69 / 28.90 \\ \hline
\end{tabular}
}
\vspace{-6mm}
\end{table*}

We follow a four-step standardized data annotation pipeline: image preprocessing, symbol standardization, character standardization, and parsing standardization. The entire annotation process took six months. For character standardization, we conduct character-by-character verification using established paleographic reference tools---including Yinqi Wenyuan\footnote{Yinqi Wenyuan (\begin{CJK}{UTF8}{gkai}殷契文渊\end{CJK}): \url{https://jgw.aynu.edu.cn}}, Guyin Xiaojing\footnote{Guyin Xiaojing (\begin{CJK}{UTF8}{gkai}古音小镜\end{CJK}): \url{http://www.kaom.net}}, Zitong Wang\footnote{Zitong Wang (\begin{CJK}{UTF8}{gkai}字统网\end{CJK}): \url{https://zi.tools}}, and Shuowen Jiezi\footnote{Shuowen Jiezi (\begin{CJK}{UTF8}{gkai}说文解字\end{CJK}): \url{http://www.shuowen.net}}---to ensure close visual correspondence between the annotation and the original glyph forms. The dataset was annotated by 20 trained annotators, refined based on feedback from domain experts in Chinese language and history, and subsequently double-checked by an independent reviewer to ensure annotation quality. Detailed tool information and annotation workflow are provided in Appendices \ref{sec:tools} and \ref{sec:annotation_flow}.

\vspace{-1mm}
\section{Experiment}
\vspace{-2mm}

We comprehensively evaluated general vision--language models and OCR-specialist models. We report \textbf{F1} \cite{ccocr} and \textbf{Normalized Edit Distance (NED)}, both computed at the character level, where
\[
\mathrm{NED}=1-\frac{D(s_{\mathrm{pred}},\,s_{\mathrm{gt}})}{\max\left(|s_{\mathrm{pred}}|,\,|s_{\mathrm{gt}}|\right)}.
\]
Here \(D(\cdot,\cdot)\) stands for the Levenshtein distance, \(s_{\mathrm{pred}}\) denotes the predicted text, \(s_{\mathrm{gt}}\) denotes the corresponding ground truth, and \(|\cdot|\) denotes string length. Prompt details can be found in Appendix~\ref{sec:prompt}.

\vspace{-2mm}
\subsection{Evaluation results on Ancient-Bench}
\subsubsection{Quantitative Experiments}
Table~\ref{tab: statistics} presents the full evaluation results. Overall, all models perform poorly on Ancient-Bench: even the best-performing model Doubao-seed-2-0-lite-260428 ( \citealp{doubao}) achieves only 52.08\% NED and 57.86\% F1, underscoring the difficulty of ancient artifact recognition. General VLMs outperform OCR-specific VLMs, reflecting the limitations of domain-specific OCR systems when confronted with heterogeneous historical scripts and media. Among all models, kimi-k2.6 achieves state-of-the-art performance. Within the OCR-specific VLM category, HunyuanOCR demonstrates potential, attaining 40.42\% NED and 46.11\% F1---the strongest result among specialist models. Furthermore, pipeline-based OCR-specific VLMs consistently outperform end-to-end OCR models, suggesting that modular recognition pipelines are better suited to the structural diversity and degradation patterns of ancient artifacts.

\vspace{-1mm}
\subsubsection{Qualitative Experiments}
\noindent \textbf{Qualitative Analysis of Medium-Specific Errors}
\noindent \textbf{(1) Oracle and Bronze.} Most models perform poorly on both tasks. The challenge in these tasks lies in the accurate transcription of ancient scripts into corresponding modern Chinese characters. For the Oracle task, Gemini achieves the best performance, but with an F1 of only 15.75\%. Visualization results reveal that models tend to produce visually similar simplified characters. For the Bronze task, Doubao-seed-2-0-lite-260428 achieves 38.48\% F1; however, visualization results show that models fail to identify complex bronze inscription glyph forms. OCR-specific VLMs completely fail on both tasks.

\begin{figure*}[!ht]
  \includegraphics[width=0.95\linewidth]{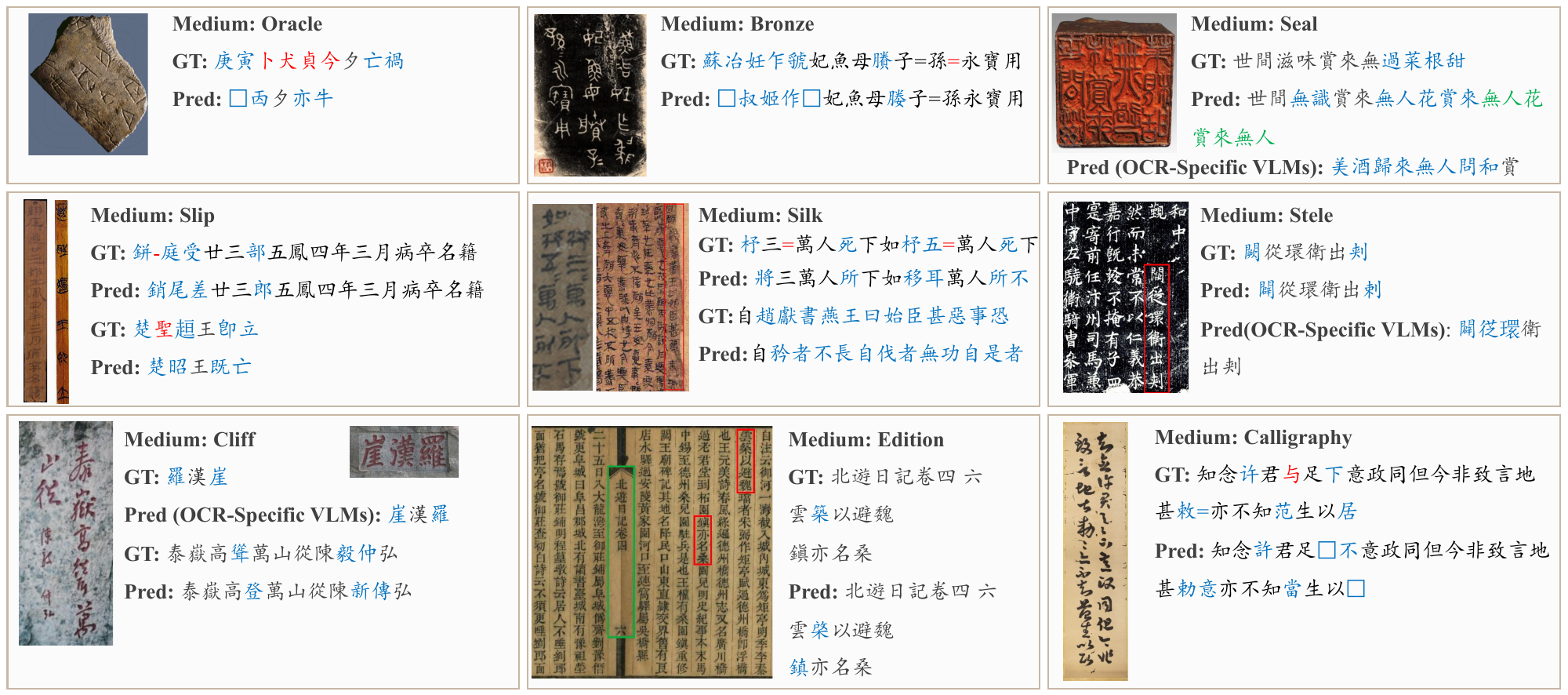}
  \centering
  \vspace{-2mm}
    \caption{\textbf{Qualitative visualization of medium-specific errors}. Red, Green, and Blue indicate \textcolor{red}{deletions}, \textcolor[RGB]{0,176,80}{insertions}, and \textcolor[RGB]{0,112,192}{substitutions}, respectively.}

  \label{fig: error_examples}
  \vspace{-2mm}
\end{figure*}

\begin{figure*}[!ht]
  \includegraphics[width=0.95\linewidth]{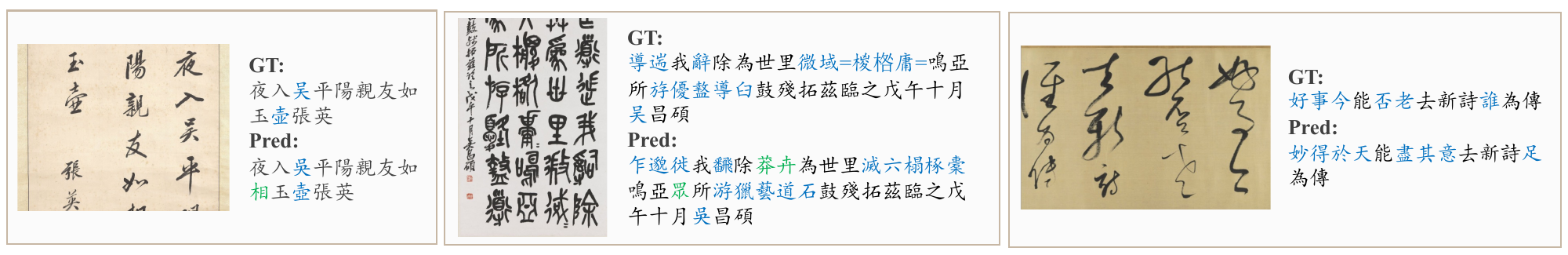}
  \centering
  \vspace{-2mm}
    \caption{\textbf{Qualitative visualization of hallucinations in general VLMs on the calligraphy subset.} Red, Green, and Blue indicate \textcolor{red}{deletions}, \textcolor[RGB]{0,176,80}{insertions}, and \textcolor[RGB]{0,112,192}{substitutions}, respectively.}
  \label{fig: error_examples_2}
  \vspace{-2mm}
\end{figure*}

\noindent \textbf{(2) Slip and Silk}. Slips are elongated text regions with low aspect ratios. The highest scores, at 39.14\% and 52.21\%, respectively. Recognition errors are caused by confusion among visually similar characters, along with failures to recognize complex archaic characters, rare characters, and special symbols. For damaged or worn regions, Silk suffers from hallucinated outputs.

\noindent \textbf{(3) Seal}. Seals require recognition of mirrored and relief features. OCR-specific VLMs struggle with this task, achieving an F1 of 18.59\%, while general VLMs reach 44.56\%. Since seal inscriptions commonly contain classical poetry, general VLMs suffer from hallucinated outputs.

\noindent \textbf{(4) Stele}. Doubao-seed-2-0-lite achieves the best result, with 88.88\%. The main difficulty lies in recognizing rare characters.

\noindent \textbf{(5) Cliff and Edition}. Both tasks present reading-order challenges, with Cliff arising from natural scenes and Edition from layout design. ocr-specific VLMs show a substantial 6--10 point gap between NED and F1 scores. In the Cliff task, general VLMs mainly struggle with cursive script, and OCR-specific VLMs frequently produce reading-order errors. In the Edition task, the main challenges include rare character recognition and layout-induced reading-order errors.

\noindent \textbf{(6) Calligraphy}. Most models perform poorly on cursive and running scripts. Additionally, distinctive symbols in calligraphic works are frequently missed or misrecognized.

\noindent \textbf{Qualitative Analysis of Hallucination Errors}

We construct a calligraphy recognition subset to evaluate hallucinations in general VLMs. As shown in Figure \ref{fig: error_examples_2}, we observe two dominant patterns: (1) \emph{prior-driven semantic completion}, where language priors under complex styles (e.g., stroke adhesion and deformation) yield fluent but non-grounded tokens (substitutions, insertions, and deletions); and (2) \emph{cropping/detection-induced over-recognition}, where inaccurate crops or boxes include neighboring strokes or noise, causing spurious extra characters and concatenated readings.


\vspace{-3mm}
\section{Conclusion}
\vspace{-2mm}

We present \textbf{Ancient-Bench}, a benchmark for ancient Chinese artifact text recognition, with \textbf{2,700} annotated images \textbf{spanning 3,000 years, 9 media types, and 7 script forms}. To mitigate dataset fragmentation, we define three unified annotation standards: symbol standardization, character standardization, and parsing standardization. Evaluations on general VLMs and OCR-specialist systems show that the ancient Chinese artifact text recognition task remains unsolved: the best model achieves only 52.08\% NED and 57.86\% F1, with particularly weak performance on oracle bones and bronze inscriptions. We further identify common failure modes, including variant/rare character confusion, missed symbols, layout-induced reading order errors, and hallucinations under visual ambiguity, highlighting Ancient-Bench as a challenging and practical benchmark for cultural heritage digitization.

\section*{Limitations}
Ancient-Bench can be further strengthened in future iterations by adding character-level bounding boxes. While the current dataset provides line-level or region-level transcriptions---effective for evaluating end-to-end recognition models and MLLMs---character-level spatial annotations would better support fine-grained text detection evaluation, and we plan to enrich them in future updates. A separate limitation is potential data contamination: since our data are sourced from publicly available repositories, we cannot guarantee that existing open-source or closed-source models have not been trained on portions of these materials.

\section*{Ethical Statement}

Ancient-Bench is built from images published by major cultural heritage institutions and other publicly accessible sources. We do not claim or transfer any copyright in the images as part of this project. Users must strictly comply with the applicable licenses and terms of use. This benchmark is intended for academic research use only.

\section*{Acknowledgement }

This research is supported in part by the National Natural Science Foundation of China (Grant No. 62476093), the Natural Science Foundation of Guangdong Province (Grant No. 2026A1515012038), the China Postdoctoral Science Foundation (Grant No. 2026M791625), and the Postdoctoral Fellowship Program (Grade B) of the China Postdoctoral Science Foundation (Grant No. GZB20260386).


\newpage

\bibliography{custom}

\clearpage
\newpage

\appendix
\section{Appendices}
\label{sec:appendix}

\subsection{Medium and Chinese Script Categories}
\label{sec:appendix_A1}

Ancient-Bench constructs a continuous temporal trajectory spanning over 3,000 years of ancient Chinese texts, organized into three historical periods. The medium types and script systems in each period are detailed below.

\noindent \textbf{Early Civilization Period (1200 - 221 BCE)}

\noindent \textbf{Oracle Bones (Oracle)}: Late Shang to early Western Zhou; used for royal divination; presented in pictographic, incised forms; characterized by surface cracks, variant characters, and ligatures.

\noindent \textbf{Bronzes}: Shang and Zhou periods, primarily used for ritual-vessel inscription engravings. Rubbings exhibit ink-bleed effects, flying-white traces, and discontinuous or connected strokes, with inscriptions interwoven with zoomorphic and cloud-thunder patterns.

\noindent \textbf{Bamboo/Wooden Slips (Slip)}: Spring and Autumn to Han periods, recording official documents and works from various philosophical schools. Bamboo and wooden slips are used as writing supports, with scripts primarily in archaic seal and clerical forms. Material deterioration causes ink fading, slip fragmentation, and text blurring.

\noindent \textbf{Silk Manuscripts (Silk)}: Mid-to-late Warring States to Han periods, using silk fabric as the writing medium. As an expensive and thin material, silk is prone to damage and was primarily used to transcribe treasured classics, divination texts, and other elite literature, during the key transition from seal to clerical scripts.

\noindent \textbf{Imperial Medieval Period and Golden Age of Stone Inscriptions (221 BCE - 900 CE)}

\noindent \textbf{Seal}: Qin--Han to Sui--Tang periods, with scripts evolving from Qin seal script to more practical (twisted) seal styles, and also including oracle-bone and bronze-inscription forms. Seals were primarily used as official credentials, for personal identity verification, document sealing, and authentication marks for calligraphy and painting collections.

\noindent \textbf{Steles}: Flourished from Han--Wei to Sui--Tang periods, encompassing tomb epitaphs from various dynasties, land-purchase contracts, and stone classics. As core physical exemplars of the evolution and standardization from clerical to regular script, steles are mostly preserved and transmitted in the form of rubbings. They were primarily used to record the deceased's life history, clan lineage, and lifetime achievements. Scripts include seal, clerical, regular, and running.

\noindent \textbf{Cliff Inscriptions (Cliff):} Han to Tang--Song periods, with the practice continuing through subsequent dynasties to the present day. These texts are carved on natural cliff faces and undergo prolonged weathering and erosion; combined with rock textures, this results in blurred characters and fragmented strokes. Scripts are predominantly clerical and regular, with occasional running and seal styles.

\noindent \textbf{Early Modern Printing and Artistic Maturity Period (900 CE - 1911 CE)}

\noindent \textbf{Ancient Chinese Editions (Edition):} Primarily consisting of woodblock-printed texts, flourishing during the Song, Yuan, Ming, and Qing dynasties. Woodblock editions feature mature and standardized layout conventions, including typical structures such as centerfold strips and interlinear notes, and serve as core materials for contemporary digitization and textual restoration of ancient books. Scripts include seal, clerical, regular, cursive, and running.

\noindent \textbf{Calligraphy}: Centered on ink-on-paper calligraphic works from the Ming and Qing dynasties, encompassing forms such as private correspondence, poetry drafts, and inscriptions on paintings and calligraphy. These works span multiple script styles, with cursive script being particularly distinctive. Some cursive character forms deviate substantially from their regular-script prototypes, requiring knowledge of calligraphic traditions and contextual understanding for accurate identification.

\subsection{Data Annotation}

\subsection{Professional Paleographic Reference Tools}
\vspace{-1mm}
\label{sec:tools}

\noindent \textbf{Yinqi Wenyuan} (Oracle Bone Script)\footnote{Yinqi Wenyuan (\begin{CJK}{UTF8}{gkai}殷契文渊\end{CJK}): \url{https://jgw.aynu.edu.cn}}: It is a non-profit oracle-bone big-data platform jointly built by the Oracle Bone Inscriptions Information Processing Key Laboratory (Anyang Normal University) and the Oracle Bone Studies and Shang History Research Center (Chinese Academy of Social Sciences). It integrates an oracle-bone rubbing/image collection, a glyph-form database, and a literature repository, and provides search and cross-database linking to support retrieval and analysis.

\noindent \textbf{Guyin Xiaojing} (Oracle/Bronze/Silk Scripts)\footnote{Guyin Xiaojing (\begin{CJK}{UTF8}{gkai}古音小镜\end{CJK}): \url{http://www.kaom.net}}: A shared platform for historical linguistics and ancient paleography. It provides paleographic glyph databases (e.g., oracle bone, bronze, and Chu-slip scripts) as well as tools for Chinese historical phonology (especially Old Chinese), dialect comparison, and queries related to phonetic components and character loans, supporting glyph lookup and comparison for research and analysis.

\noindent \textbf{Zitong Wang} (Multi-Script Coverage)\footnote{Zitong Wang (\begin{CJK}{UTF8}{gkai}字统网\end{CJK}): \url{https://zi.tools}}: A comprehensive Chinese character information platform containing tens of thousands of characters, spanning oracle bone script, bronze inscriptions, seal script, and clerical script to modern simplified and traditional forms. It provides glyph evolution, etymological decomposition, pronunciation variants, and encoding queries, with extensive coverage of rare and variant characters.

\noindent \textbf{Shuowen Jiezi} (Multi-Script Coverage)\footnote{Shuowen Jiezi (\begin{CJK}{UTF8}{gkai}说文解字\end{CJK}): \url{http://www.shuowen.net}}: An online reference built around \emph{Shuowen Jiezi}, enabling radical-based character lookup and providing entries with traditional explanations and later annotations, which is helpful for paleographic comparison and etymological research in classical Chinese.

\subsection{Example Annotation Workflow}
\label{sec:annotation_flow}
Taking the oracle-bone medium subset as an example, we first locate the corresponding rubbings in professional oracle-bone inscription tools such as Yinqi Wenyuan (\begin{CJK}{UTF8}{gkai}殷契文渊\end{CJK}) based on information from the original artifacts. We then cross-check against the transcriptions in ``Collected Transcriptions and Interpretations'' (\begin{CJK}{UTF8}{gkai}摹释总集释文\end{CJK}) and conduct character-by-character verification via Chinese character search in the glyph database, ensuring that each glyph precisely matches its interpretation.

To further improve data accuracy, we additionally use another specialized tool, Guyin Xiaojing (\begin{CJK}{UTF8}{gkai}古音小镜\end{CJK}), to validate individual characters and perform a second-round review. Overall, our annotation pipeline strictly follows a workflow of ``cross-validation across multiple tools + multiple rounds of annotator verification + final approval,'' with rigorous quality control throughout.

\begin{figure}[t]
  \centering
  \includegraphics[width=\linewidth]{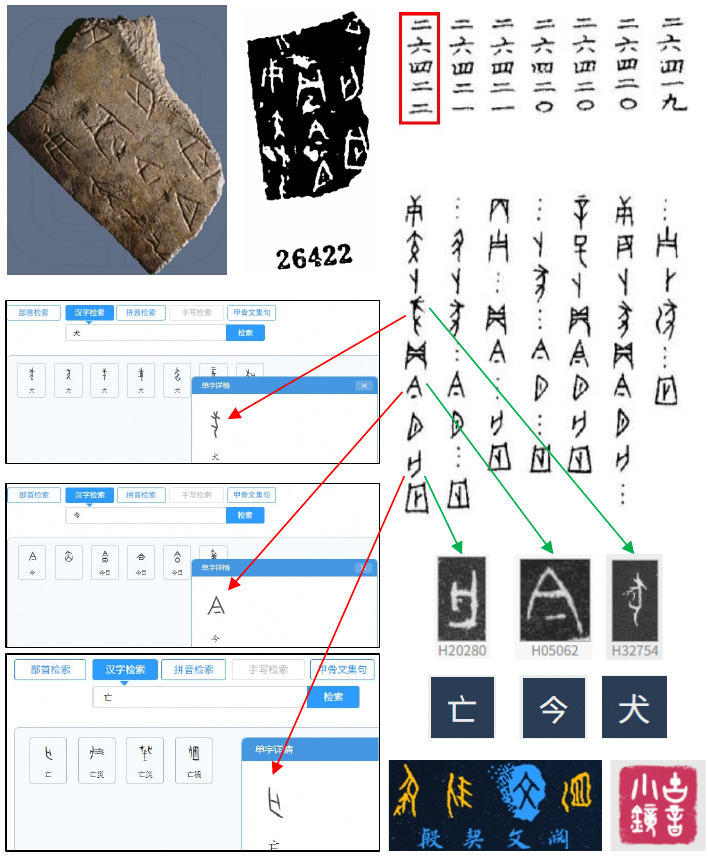}
  \vspace{-6mm}
  \caption{Oracle-Bone Example Annotation Workflow}
  \label{fig: annotation_flow}
  \vspace{-7mm}
\end{figure}

\subsection{Prompt Details}
\label{sec:prompt}

\paragraph{Base Rule Prompt.}
To handle the distinctive characteristics of ancient Chinese texts (e.g., special symbols, reading order, and damaged characters), we design a Base Rule Prompt that explicitly specifies nine constraints for the recognition task.

Ancient artifacts across different media exhibit substantial variation in physical condition, layout structure, and writing conventions, making a single unified prompt insufficient for all document types. For example, woodblock editions often contain complex layout elements such as interlinear notes and page margins. To better guide MLLMs and mitigate these domain-specific issues, we design task-specific prompts.

In the following sections, we detail the task-specific prompts for each document type.

\subsection{Evaluation Metrics}
We apologize for the confusion and clarify the evaluation protocol here. To reduce unfairness caused by formatting noise that is irrelevant to the content, we apply a unified normalization procedure to both the predicted text and the ground truth (GT) before computing the metrics. The details are as follows:

\begin{itemize}[\textbullet]
    \item \textbf{Special tags / placeholders} (\texttt{\textless undeciphered\textgreater}, \texttt{\textless unrecognizable\textgreater}, \begin{CJK}{UTF8}{gkai}□\end{CJK}): during tokenization and alignment, we treat them as indivisible atomic tokens to prevent them from being split into multiple characters and receiving unreasonable penalties.
    \item \textbf{Counting repeated symbols / repeated characters}: for repeatable placeholder tokens, we use a count-based matching strategy. Using \begin{CJK}{UTF8}{gkai}□\end{CJK} as an example, if it appears 3 times in the GT but only once in the prediction, then the matching contribution of this token is $\min(3, 1) = 1$.
    \item \textbf{Whitespace handling}: outputs from OCR / multimodal models often include automatic line breaks or extra spaces. We remove all whitespace characters---including spaces, newlines, tabs, etc.---to further reduce the impact of line breaks/formatting on scoring.
   \item \textbf{Simplified vs. Traditional handling}: we do \textbf{not} perform Simplified/Traditional Chinese normalization. In classical books and rubbings, simplified / traditional / variant-form differences are themselves part of the recognition difficulty; unifying them may obscure genuine model differences.
    \item \textbf{Punctuation handling}: in edit-distance-based similarity computation (as well as hallucination-rate statistics), we ignore both Chinese and English punctuation marks (e.g., \begin{CJK}{UTF8}{gkai}“。” “，” “；” “：” “？” “！” “、”\end{CJK}, etc.) to reduce the impact of inconsistent punctuation annotation.
    \item \textbf{Case handling}: we convert all English letters to lowercase.
\end{itemize}


\begin{figure*}[!htbp]
\centering
\begin{tcolorbox}[colback=gray!5!white, colframe=gray!75!black, title=\textbf{Base Rule Prompt (Chinese)}, arc=2mm]
\ttfamily
\begin{CJK}{UTF8}{gkai}
请帮我识别图片中的文字，注意：\\
（1）剔除标点： 不要输出任何现代句读（标点符号）；\\
（2）保留古符号： 必须保留原文中出现的特定古文形符（重文号“=”、“-”，交换符号“\textasciitilde{}”，三点删除符统一用"、"表示）；\\
（3）残损占位： 遇到完全模糊或破损无法识别的字，请使用 □ 代替，切勿自行编造；\\
（4）换行规则： 图片中的文字每一垂直列为一行，请按列换行输出；\\
（5）所见即所得：请严格保留图片中实际书写的原始繁体字、生僻字和异体字形，绝对不可自动转换为现代通行繁体字或简体字；\\
（6）阅读顺序：请严格按照人类实际阅读该文献的顺序输出。通常仅为“从右到左”或“从左到右”逐列推进，不存在各列之间跳跃交错阅读的情况；\\
（7）除了纯印章数据以外的数据如果存在印章，均不识别；\\
（8）严格可见范围：只输出图片内真正可见的文字，绝对不可利用先验知识补全图片中未出现的文字内容；\\
（9）输出格式：只输出识别结果，不要包含任何多余的对话、问候或解释性文字。换行采用"\textbackslash n"符号，请将最终的纯文本严格包裹在 <res> 和 </res> 标签内部。
\end{CJK}
\end{tcolorbox}
\end{figure*}

\begin{figure*}[t]
\centering
\begin{tcolorbox}[colback=gray!5!white, colframe=gray!75!black, title=\textbf{Base Rule Prompt (English)}, arc=2mm]
\ttfamily
Please help me recognize the text in the image. Note:\\
(1) Remove punctuation: Do not output any modern punctuation marks;\\
(2) Retain ancient symbols: You must retain specific ancient typographic symbols appearing in the original text (repetition marks "=" or "-", transposition mark "\textasciitilde{}", and three-dot deletion marks uniformly represented by "\begin{CJK}{UTF8}{gkai}、\end{CJK}");\\
(3) Placeholder for damage: For completely blurred or damaged characters that cannot be recognized, please use "\begin{CJK}{UTF8}{gkai}□\end{CJK}" as a placeholder. Do not fabricate characters;\\
(4) Line break rule: Each vertical column of text in the image corresponds to one line. Please output with line breaks according to the columns;\\
(5) What you see is what you get: Strictly retain the original traditional characters, rare characters, and variant forms actually written in the image. Absolutely do not automatically convert them to modern standard traditional or simplified characters;\\
(6) Reading order: Strictly output in the actual human reading order of the document. This is typically column-by-column from right to left or left to right, without jumping or alternating between columns;\\
(7) Ignore seals: Unless the data consists purely of a seal, do not recognize any seal text present in the image;\\
(8) Strict visibility scope: Only output text that is genuinely visible within the image. Absolutely do not use prior knowledge to complete text that does not appear in the image;\\
(9) Output format: Only output the recognition results. Do not include any extra dialogue, greetings, or explanatory text. Use "\textbackslash n" for line breaks, and strictly enclose the final plain text within <res> and </res> tags.
\end{tcolorbox}
\end{figure*}

\begin{figure*}[t] 
\centering
\begin{tcolorbox}[colback=gray!5!white, colframe=gray!75!black, title=\textbf{Oracle Prompt (Chinese)}, arc=2mm]
\ttfamily
\begin{CJK}{UTF8}{gkai}
这是一张甲骨实物的古籍图片，请帮我识别图片中的文字\\
\textbf{[此处插入基础规则 Prompt]}\\
一、甲骨文字隶定与未破译文字专项规范\\
（1）可破译字形处理：明确对应学界公认隶定字形的甲骨文字，依照权威甲骨释读规范，输出其对应的现代隶定文字；\\
（2）未破译文字强制规则：目前已发现甲骨单字存量大，仅少部分完成权威释读，凡是无统一公认释读结论、学界尚未破译的甲骨字形，禁止凭形近字随意猜测判定，统一固定使用 <undeciphered> 标签进行标注，不得随意替换成常用汉字；\\
（3）字形区分原则：严格区分人工契刻文字与甲骨天然纹理，只识别人为刻写卜辞内容。\\
二、甲骨实物特有干扰处理规范\\
（1）区分“兆纹”与文字： 甲骨表面存在大量因占卜产生的自然裂纹（兆纹，通常呈现为粗大的横竖交错线）。请仔细甄别，绝对不可将骨面本身的自然裂缝、钻凿痕迹误识别为汉字“一”、“二”、“卜”、“十”等。\\
三、输出格式\\
（1）只输出识别结果，不要包含任何多余的对话、问候或解释性文字。换行采用"\textbackslash n"符号，请将最终的纯文本严格包裹在 <res> 和 </res> 标签内部；\\
（2）输出示例：<res>文本内容\textbackslash n文本内容\textbackslash n文本内容<undeciphered>...</res>。例如：<res>辛亥卜贞今秋禾\textbackslash n王占曰吉\textbackslash n<undeciphered>卯祀用</res>。
\end{CJK}
\end{tcolorbox}
\end{figure*}

\begin{figure*}[t]
\centering
\begin{tcolorbox}[colback=gray!5!white, colframe=gray!75!black, title=\textbf{Bronze Inscription Prompt (Chinese)}, arc=2mm]
\ttfamily
\begin{CJK}{UTF8}{gkai}
这是一张青铜器铭文（金文）或金石拓片的古籍图片，请帮我识别图片中的文字\\
\textbf{[此处插入基础规则 Prompt]}\\
一、金文隶定与无统一释读文字处理规范\\
（1）查询金文隶定：请识别图片中的金文，依照商周金文权威释读规范，输出其对应的现代隶定汉字；\\
（2）无统一释读文字：遇到族徽文字、图形金文、异形古字，无公认隶定字形时，统一使用 <undeciphered> 标签标注，禁止形近字臆测；\\
二、金石拓片特有干扰处理规范\\
（1）区分“石花/范垢”与笔画： 拓片常有白色的斑点（石花）或青铜器铸造时的范垢阴影。请仔细甄别，绝对不可将拓片上的破损白斑、背景杂色误识别为文字笔画（如误认为“一”、“丶”等）；\\
（2）忽略器物纹饰：青铜器上常伴有饕餮纹、云雷纹等几何纹饰。请严格区分文字区与纹饰区，不要将纹饰线条识别为文字；\\
（3）合文处理：金文中有少量合文（如“子孙”合写），识别时请将其语义拆开，直接输出为两个独立的汉字。\\
三、输出格式\\
（1）只输出识别结果，不要包含任何多余的对话、问候或解释性文字。换行采用"\textbackslash n"符号，请将最终的纯文本严格包裹在 <res> 和 </res> 标签内部；\\
（2）输出示例：<res>唯正月初吉 丁亥\textbackslash n王□作其宝尊彝\textbackslash n子孙<undeciphered>宝用</res>
\end{CJK}
\end{tcolorbox}
\end{figure*}

\begin{figure*}[t]
\centering
\begin{tcolorbox}[colback=gray!5!white, colframe=gray!75!black, title=\textbf{Oracle Prompt (English)}, arc=2mm]
\ttfamily
This is an image of an original oracle bone ancient text. Please help me recognize the text in the image.\\
\textbf{[Insert Base Rule Prompt here]}\\
I. Special Specifications for Oracle Bone Character Transcription and Undeciphered Characters\\
(1) Handling decipherable characters: For oracle bone characters with clearly recognized modern counterparts accepted by academia, output their corresponding modern transcribed characters according to authoritative oracle bone decipherment standards;\\
(2) Mandatory rule for undeciphered characters: Currently, there is a large inventory of discovered oracle bone characters, but only a small portion has been authoritatively deciphered. For any oracle bone characters without a unified, universally accepted decipherment conclusion and not yet deciphered by academia, guessing based on visual similarity is prohibited. Uniformly and fixedly use the <undeciphered> tag to mark them, and do not arbitrarily replace them with common Chinese characters;\\
(3) Character differentiation principle: Strictly distinguish between artificially engraved text and the natural texture of the oracle bone. Only recognize artificially engraved divination text.\\
II. Special Interference Handling Specifications for Physical Oracle Bones\\
(1) Distinguish "divination cracks" (zhaowen) from text: There are many natural cracks (zhaowen, usually appearing as thick intersecting horizontal and vertical lines) on the surface of oracle bones caused by divination. Please identify them carefully. Absolutely do not misidentify the natural cracks and drilling/chiseling marks on the bone surface as Chinese characters like "\begin{CJK}{UTF8}{gkai}一\end{CJK}" (one), "\begin{CJK}{UTF8}{gkai}二\end{CJK}" (two), "\begin{CJK}{UTF8}{gkai}卜\end{CJK}" (divine), or "\begin{CJK}{UTF8}{gkai}十\end{CJK}" (ten).\\
III. Output Format\\
(1) Only output the recognition results. Do not include any extra dialogue, greetings, or explanatory text. Use the "\textbackslash n" symbol for line breaks, and strictly enclose the final plain text within <res> and </res> tags;\\
(2) Output example: <res>text content\textbackslash ntext content\textbackslash ntext content<undeciphered>...</res>. For example: <res>\begin{CJK}{UTF8}{gkai}辛亥卜贞今秋禾\end{CJK}\textbackslash n\begin{CJK}{UTF8}{gkai}王占曰吉\end{CJK}\textbackslash n<undeciphered>\begin{CJK}{UTF8}{gkai}卯祀用\end{CJK}</res>.
\end{tcolorbox}
\end{figure*}

\begin{figure*}[t]
\centering
\begin{tcolorbox}[colback=gray!5!white, colframe=gray!75!black, title=\textbf{Bronze Inscription Prompt (English)}, arc=2mm]
\ttfamily
This is an image of an ancient text featuring bronze inscriptions (Jinwen) or rubbings of metal and stone. Please help me recognize the text in the image.\\
\textbf{[Insert Base Rule Prompt here]}\\
I. Processing Specifications for Bronze Inscription Transcription and Undeciphered Characters\\
(1) Query bronze inscription transcription: Please recognize the bronze inscriptions in the image and output their corresponding modern transcribed Chinese characters according to the authoritative decipherment standards for Shang and Zhou bronze inscriptions;\\
(2) Undeciphered characters: When encountering clan emblem characters, pictographic bronze inscriptions, or variant ancient characters without universally accepted transcribed forms, uniformly use the <undeciphered> tag to mark them. Guessing based on visual similarity is prohibited;\\
II. Special Interference Handling Specifications for Rubbings of Metal and Stone\\
(1) Distinguish "stone flowers/casting dirt" from strokes: Rubbings often have white spots (stone flowers) or shadows of casting dirt from when the bronze vessel was cast. Please identify them carefully. Absolutely do not misidentify damaged white spots or background noise on the rubbing as text strokes (e.g., mistaking them for "\begin{CJK}{UTF8}{gkai}一\end{CJK}" or "\begin{CJK}{UTF8}{gkai}丶\end{CJK}");\\
(2) Ignore vessel decorations: Bronze vessels are often accompanied by geometric decorations such as Taotie patterns and cloud-and-thunder patterns. Please strictly distinguish the text area from the decoration area, and do not recognize decoration lines as text;\\
(3) Handling ligatures (Hewen): There are a few ligatures in bronze inscriptions (e.g., "\begin{CJK}{UTF8}{gkai}子孙\end{CJK}" written together). When recognizing them, please separate their semantics and directly output them as two independent Chinese characters.\\
III. Output Format\\
(1) Only output the recognition results. Do not include any extra dialogue, greetings, or explanatory text. Use the "\textbackslash n" symbol for line breaks, and strictly enclose the final plain text within <res> and </res> tags;\\
(2) Output example: <res>text content\textbackslash ntext content\textbackslash ntext content<undeciphered>...</res>. For example: <res>\begin{CJK}{UTF8}{gkai}唯正月初吉 丁亥\end{CJK}\textbackslash n\begin{CJK}{UTF8}{gkai}王□作其宝尊彝\end{CJK}\textbackslash n\begin{CJK}{UTF8}{gkai}子孙\end{CJK}<undeciphered>\begin{CJK}{UTF8}{gkai}宝用\end{CJK}</res>.
\end{tcolorbox}
\end{figure*}

\begin{figure*}[t]
\centering
\begin{tcolorbox}[colback=gray!5!white, colframe=gray!75!black, title=\textbf{Bamboo/Wooden Slip Prompt (Chinese)}, arc=2mm]
\ttfamily
\begin{CJK}{UTF8}{gkai}
这是一张简牍的古籍图片，请帮我识别图片中的文字\\
\textbf{[此处插入基础规则 Prompt]}\\
一、生僻字，异体字\\
（1）常规字形严格保留图片中实际书写的原始繁体字、生僻字和异体字形，绝对不可自动转换为现代通行繁体字或简体字；\\
（2）特殊情况，简牍中常存在计算机 Unicode 字库尚未编码的古体字/生僻字\\
\quad（a）优先替换： 若该未编码字存在可打出的异体字或对应通行字（例如：图片上是从上到下结构为“虎口虫虫”的字，字库无此字，但其对应异体字为“虐”），则允许使用该异体字“虐”替代输出；\\
\quad（b）无法替换时占位： 若该未编码字既无法打出，也完全找不到任何对应的异体字/通行字，请严格使用 <unrecognizable> 标签代替该位置。\\
二、简牍特有符号与版面元素处理规范\\
（1）特殊符号（原样保留）：在简牍中，符号“=”的含义极为复杂（可能表示重文、合文、专名号或字形省略等）。为了保留原始文献信息供专家后续考证，遇到“=”或“-”等符号时，请绝对不要自行展开或解释，必须原样输出该符号；\\
（2）合文现象（拆分输出）：简牍中常将两个字拼写为一个整体（如将“五十”、“大夫”、“之子”等字上下连写或共用偏旁）。当你识别到这种视觉上的“合文”时，请将其语义拆开，直接输出为两个独立的汉字；\\
（3）简牍残断与留白：简牍顶部和底部常有物理断裂。如果在行中或头尾存在明显的字迹剥落或残损坑洞，确信原本有字但已消失，请输出 □；如果是古人原本就未写字的空白竹简区域，则无需输出任何符号；\\
（4）行内空格：如果同一列简牍中，上下文字之间存在明显的物理留白（如段落切分、句读停顿或抬头），请在对应的文字之间插入一个空格符号“ ”，自然嵌入到它所属的正文位置中。\\
三、输出格式\\
（1）只输出识别结果，不要包含任何多余的对话、问候或解释性文字。换行采用"\textbackslash n"符号，请将最终的纯文本严格包裹在 <res> 和 </res> 标签内部；\\
（2）输出示例：<res>文字内容...文字五十正常拆分...遇到子=保留符号...遇到未编码字且无替换词用<unrecognizable>占位\textbackslash n...上半部分 下半部分\textbackslash n...文字内容</res>。例如：<res>大夫之子=皆至\textbackslash n五十有六年 天下大旱\textbackslash n臣□奉<unrecognizable>死罪死罪</res>
\end{CJK}
\end{tcolorbox}
\end{figure*}

\begin{figure*}[t]
\centering
\begin{tcolorbox}[colback=gray!5!white, colframe=gray!75!black, title=\textbf{Bamboo/Wooden Slip Prompt (English)}, arc=2mm]
\ttfamily
This is an image of an ancient text on bamboo or wooden slips (Jiandu). Please help me recognize the text in the image.\\
\textbf{[Insert Base Rule Prompt here]}\\
I. Rare and Variant Characters\\
(1) For regular character forms, strictly retain the original traditional Chinese characters, rare characters, and variant character forms actually written in the image. Absolutely do not automatically convert them into modern standard traditional or simplified characters;\\
(2) Special circumstances: Bamboo and wooden slips often contain ancient/rare characters that have not yet been encoded in the computer Unicode font library.\\
\quad(a) Priority replacement: If the unencoded character has a typable variant or corresponding standard character (for example, if the image shows a character with a top-to-bottom structure of "\begin{CJK}{UTF8}{gkai}虎口虫虫\end{CJK}", which is not in the font library, but its corresponding variant is "\begin{CJK}{UTF8}{gkai}虐\end{CJK}"), you are allowed to use the variant "\begin{CJK}{UTF8}{gkai}虐\end{CJK}" as a substitute in the output;\\
\quad(b) Placeholder when irreplaceable: If the unencoded character cannot be typed and no corresponding variant/standard character can be found at all, strictly use the <unrecognizable> tag to replace that position.\\
II. Processing Specifications for Slip-Specific Symbols and Layout Elements\\
(1) Special symbols (retain as is): In bamboo and wooden slips, the meaning of the symbol "=" is extremely complex (it may indicate repetition, ligature, proper noun mark, or character omission, etc.). To preserve original document information for subsequent expert textual research, when encountering symbols like "=" or "-", absolutely do not expand or explain them yourself; you must output the symbol exactly as it is;\\
(2) Ligature phenomenon (split output): In slips, two characters are often spelled together as a whole (e.g., writing "\begin{CJK}{UTF8}{gkai}五十\end{CJK}", "\begin{CJK}{UTF8}{gkai}大夫\end{CJK}", "\begin{CJK}{UTF8}{gkai}之子\end{CJK}" vertically connected or sharing components). When you recognize this visual "ligature", please separate their semantics and directly output them as two independent Chinese characters;\\
(3) Slip breakage and blank space: The top and bottom of slips often have physical breaks. If there is obvious ink peeling or a damaged hole in the middle, beginning, or end of a line, and you are certain there was originally a character but it has disappeared, please output \begin{CJK}{UTF8}{gkai}□\end{CJK}; if it is a blank bamboo slip area where the ancients originally wrote nothing, there is no need to output any symbols;\\
(4) Inline spaces: If there is obvious physical blank space between upper and lower text in the same column of a slip (e.g., paragraph segmentation, punctuation pause, or indentation), please insert a space symbol " " between the corresponding text, naturally embedding it into its original text position.\\
III. Output Format\\
(1) Only output the recognition results. Do not include any extra dialogue, greetings, or explanatory text. Use the "\textbackslash n" symbol for line breaks, and strictly enclose the final plain text within <res> and </res> tags;\\
(2) Output example: <res>text content... text \begin{CJK}{UTF8}{gkai}五十\end{CJK} split normally... encounter \begin{CJK}{UTF8}{gkai}子\end{CJK}= retain symbol... encounter unencoded character with no replacement word use <unrecognizable> placeholder\textbackslash n... upper part lower part\textbackslash n... text content</res>. For example: <res>\begin{CJK}{UTF8}{gkai}大夫之子=皆至\end{CJK}\textbackslash n\begin{CJK}{UTF8}{gkai}五十有六年 天下大旱\end{CJK}\textbackslash n\begin{CJK}{UTF8}{gkai}臣□奉\end{CJK}<unrecognizable>\begin{CJK}{UTF8}{gkai}死罪死罪\end{CJK}</res>
\end{tcolorbox}
\end{figure*}

\begin{figure*}[t]
\centering
\begin{tcolorbox}[colback=gray!5!white, colframe=gray!75!black, title=\textbf{Silk Manuscript Prompt (Chinese)}, arc=2mm]
\ttfamily
\begin{CJK}{UTF8}{gkai}
这是一张帛书的古籍图片，请帮我识别图片中的文字\\
\textbf{[此处插入基础规则 Prompt]}\\
一、生僻字，异体字\\
（1）常规字形严格保留图片中实际书写的原始繁体字、生僻字和异体字形，不可自动转换为现代通行繁体字或简体字；\\
（2）特殊情况，帛书中常存在计算机 Unicode 字库尚未编码的古体字/生僻字\\
\quad（a）优先替换： 若该未编码字存在可打出的异体字或对应通行字（例如：图片上是从上到下结构为“虎口虫虫”的字，字库无此字，但其对应异体字为“虐”），则允许使用该异体字“虐”替代输出；\\
\quad（b）无法替换时占位： 若该未编码字既无法打出，也完全找不到任何对应的异体字/通行字，请严格使用 <unrecognizable> 标签代替该位置。\\
二、帛书特有符号与物理干扰处理规范\\
（1）特殊符号（原样保留）：在帛书中，符号“=”的含义极为复杂（可能表示重文、合文、专名号或字形省略等）。为了保留原始文献信息供专家后续考证，遇到“=”或“-”等符号时，请绝对不要自行展开或解释，必须原样输出该符号；\\
（2）合文现象（拆分输出）：帛书中常将两个字拼写为一个整体（如将“五十”、“大夫”、“之子”等字上下连写或共用偏旁）。当你识别到这种视觉上的“合文”时，请将其语义拆开，直接输出为两个独立的汉字；\\
（3）行内空格：如果同一列中，上下文字之间存在明显的物理留白（如段落切分、句读停顿），请在对应的文字之间插入一个空格符号；\\
（4）破损现象：帛书为丝织品，易大面积腐朽、褪色、连片损毁，无法精准判定缺失字数，整片连续残缺区域统一仅用一个残损占位"□"表示，不重复叠加；\\
（5）过滤反印文与水渍：帛书因折叠存放，常出现“反印文”（相邻层墨迹渗透过来的镜像反向字迹）或严重的水渍渗墨。请严格只识别正面的、笔画清晰的主体文字，必须忽略背面透出或反向印染的模糊镜像字迹。\\
三、输出示例\\
输出示例：<res>文字内容...\textbackslash n文字五十正常拆分...遇到子=保留符号...遇到未编码字且无替换词用<unrecognizable>占位\textbackslash n...上半部分 下半部分\textbackslash n...文字内容\textbackslash n...文字内容□...文字内容</res>。例如：<res>天地不仁 以万物为刍狗\textbackslash n圣人不仁□以百姓为<unrecognizable>狗\textbackslash n道可道也=非恒道也</res>
\end{CJK}
\end{tcolorbox}
\end{figure*}

\begin{figure*}[t]
\centering
\begin{tcolorbox}[colback=gray!5!white, colframe=gray!75!black, title=\textbf{Silk Manuscript Prompt (English)}, arc=2mm]
\ttfamily
This is an image of an ancient text on silk (Boshu). Please help me recognize the text in the image.\\
\textbf{[Insert Base Rule Prompt here]}\\
I. Rare and Variant Characters\\
(1) For regular character forms, strictly retain the original traditional Chinese characters, rare characters, and variant character forms actually written in the image. Do not automatically convert them into modern standard traditional or simplified characters;\\
(2) Special circumstances: Silk manuscripts often contain ancient/rare characters that have not yet been encoded in the computer Unicode font library.\\
\quad(a) Priority replacement: If the unencoded character has a typable variant or corresponding standard character (for example, if the image shows a character with a top-to-bottom structure of "\begin{CJK}{UTF8}{gkai}虎口虫虫\end{CJK}", which is not in the font library, but its corresponding variant is "\begin{CJK}{UTF8}{gkai}虐\end{CJK}"), you are allowed to use the variant "\begin{CJK}{UTF8}{gkai}虐\end{CJK}" as a substitute in the output;\\
\quad(b) Placeholder when irreplaceable: If the unencoded character cannot be typed and no corresponding variant/standard character can be found at all, strictly use the <unrecognizable> tag to replace that position.\\
II. Processing Specifications for Silk-Specific Symbols and Physical Interferences\\
(1) Special symbols (retain as is): In silk manuscripts, the meaning of the symbol "=" is extremely complex (it may indicate repetition, ligature, proper noun mark, or character omission, etc.). To preserve original document information for subsequent expert textual research, when encountering symbols like "=" or "-", absolutely do not expand or explain them yourself; you must output the symbol exactly as it is;\\
(2) Ligature phenomenon (split output): In silk manuscripts, two characters are often spelled together as a whole (e.g., writing "\begin{CJK}{UTF8}{gkai}五十\end{CJK}", "\begin{CJK}{UTF8}{gkai}大夫\end{CJK}", "\begin{CJK}{UTF8}{gkai}之子\end{CJK}" vertically connected or sharing components). When you recognize this visual "ligature", please separate their semantics and directly output them as two independent Chinese characters;\\
(3) Inline spaces: If there is obvious physical blank space between upper and lower text in the same column (e.g., paragraph segmentation, punctuation pause), please insert a space symbol between the corresponding text;\\
(4) Damage phenomenon: Silk manuscripts are silk fabrics, prone to large-area decay, fading, and contiguous damage. It is impossible to accurately determine the number of missing characters. A continuous damaged area should be uniformly represented by only one damage placeholder "\begin{CJK}{UTF8}{gkai}□\end{CJK}", without repeated stacking;\\
(5) Filter reverse imprints and water stains: Because silk manuscripts are stored folded, "reverse imprints" (mirror-reversed handwriting permeated from adjacent layers of ink) or severe water stain ink seepage often appear. Please strictly recognize only the main text on the front with clear strokes, and you must ignore the blurred mirror handwriting showing through from the back or reversely dyed.\\
III. Output Example\\
Output example: <res>text content...\textbackslash n text \begin{CJK}{UTF8}{gkai}五十\end{CJK} split normally... encounter \begin{CJK}{UTF8}{gkai}子\end{CJK}= retain symbol... encounter unencoded character with no replacement word use <unrecognizable> placeholder\textbackslash n... upper part lower part\textbackslash n... text content\textbackslash n... text content\begin{CJK}{UTF8}{gkai}□\end{CJK}... text content</res>. For example: <res>\begin{CJK}{UTF8}{gkai}天地不仁 以万物为刍狗\end{CJK}\textbackslash n\begin{CJK}{UTF8}{gkai}圣人不仁□以百姓为\end{CJK}<unrecognizable>\begin{CJK}{UTF8}{gkai}狗\end{CJK}\textbackslash n\begin{CJK}{UTF8}{gkai}道可道也=非恒道也\end{CJK}</res>
\end{tcolorbox}
\end{figure*}

\begin{figure*}[t]
\centering
\begin{tcolorbox}[colback=gray!5!white, colframe=gray!75!black, title=\textbf{Seal Prompt (Chinese)}, arc=2mm]
\ttfamily
\begin{CJK}{UTF8}{gkai}
这是一张印章实物或印蜕图片，请帮我识别图片中的文字\\
\textbf{[此处插入基础规则 Prompt]}\\
一、字形识别与书体区分规范\\
（1）多书体识别： 印章文字虽以篆书（大篆、小篆、缪篆）为主，但也常包含甲骨文、金文（大篆）等古老书体。识别时请注意，如果是甲骨文，请明确对应学界公认隶定字形的甲骨文字，依照权威甲骨释读规范，输出其对应的现代隶定文字；如果是金文，请依照商周金文权威释读规范，输出其对应的现代隶定汉字；\\
二、印章特有识别规范\\
（1）朱文与白文： 请准确识别朱文（阳文，红字白底）和白文（阴文，白字红底），不要因色彩反转干扰识别；\\
（2）实物载体（纯石头印章）： 对于直接拍摄的石质、玉质或金属实物印章，文字多为直接雕刻且无墨迹填充。请通过材质表面的凹凸、光影和刻痕深度来辨识文字，不要将其误认为背景纹理；\\
（3）镜像处理（实物特有）： 如果图片是印章的实物底面（字是反的），请在识别时自动进行镜像翻转，输出正常的隶定汉字。如果图片是盖出的印蜕（字是正的），则直接识别。\\
三、印章排版、共用字与阅读顺序规范\\
（1）分析布局逻辑： 印章布局除常规方格形式外，常有圆印、十字印或自由分布；\\
（2）共用字处理（借笔/连读）：\\
\quad（a）识别共用： 识别印章中是否存在“中心字被多处共用”或“首尾字被多行共用”的现象；\\
\quad（b）语义重组输出： 当发现共用字时，请根据印章的设计逻辑，输出完整的语义短语，而不是机械的字迹排列。示例： 若顶部可见“不失”，底部可见“于人”，中间并列“足”、“口”、“色”，请将其重组为完整的语义短语分行输出（即：不失足于人、不失口于人、不失色于人）。\\
（3）阅读顺序与换行：\\
\quad（a）对于常规方印：按列换行（先右列后左列）；\\
\quad（b）对于共用字/十字印：请按重组后的语义短语换行输出。\\
四、输出格式\\
（1）只输出识别结果，不要包含任何多余的对话、问候或解释性文字。换行采用"\textbackslash n"符号，请将最终的纯文本严格包裹在 <res> 和 </res> 标签内部；\\
（2）输出示例：<res>文本内容...时=文本内容\textbackslash n文本内容</res>。例如：<res>晓儿之时=而分=\textbackslash n而不可再来</res>
\end{CJK}
\end{tcolorbox}
\end{figure*}

\begin{figure*}[t]
\centering
\begin{tcolorbox}[colback=gray!5!white, colframe=gray!75!black, title=\textbf{Seal Prompt (English)}, arc=2mm]
\ttfamily\fontsize{12pt}{12pt}\selectfont
This is an image of a physical seal or a seal imprint. Please help me recognize the text in the image.\\
\textbf{[Insert Base Rule Prompt here]}\\
I. Character Recognition and Script Differentiation Specifications\\
(1) Multi-script recognition: Although seal text is primarily in seal script (Large Seal, Small Seal, Miao Seal), it also frequently includes ancient scripts such as Oracle Bone Script and Bronze Script (Large Seal). When recognizing, please note: if it is Oracle Bone Script, clearly correspond it to the academically recognized clericalized (Liding) Oracle Bone characters, and output the corresponding modern clericalized text according to authoritative Oracle Bone decipherment specifications; if it is Bronze Script, output the corresponding modern clericalized Chinese characters according to authoritative Shang and Zhou Bronze Script decipherment specifications;\\
II. Seal-Specific Recognition Specifications\\
(1) Zhuwen and Baiwen: Please accurately distinguish Zhuwen (relief/yang characters, red text on white background) and Baiwen (intaglio/yin characters, white text on red background). Do not let color inversion interfere with recognition;\\
(2) Physical carrier (pure stone seals): For directly photographed physical seals made of stone, jade, or metal, the text is mostly directly carved without ink filling. Please identify the text through the unevenness, light and shadow, and carving depth of the material surface, and do not mistake it for background texture;\\
(3) Mirror processing (specific to physical objects): If the image is the bottom surface of a physical seal (the characters are mirrored), please automatically perform a mirror flip during recognition and output standard clericalized Chinese characters. If the image is a stamped seal imprint (the characters are oriented normally), recognize it directly.\\
III. Seal Layout, Shared Characters, and Reading Order Specifications\\
(1) Analyze layout logic: In addition to the regular grid format, seal layouts often include circular seals, cross-shaped seals, or free distribution;\\
(2) Shared character processing (borrowed strokes/continuous reading):\\
\quad(a) Recognize sharing: Identify whether there is a phenomenon in the seal where a "central character is shared in multiple places" or "first/last characters are shared across multiple lines";\\
\quad(b) Semantic reorganization output: When shared characters are found, please output complete semantic phrases based on the seal's design logic, rather than a mechanical arrangement of strokes. Example: If "\begin{CJK}{UTF8}{gkai}不失\end{CJK}" is visible at the top, "\begin{CJK}{UTF8}{gkai}于人\end{CJK}" at the bottom, and "\begin{CJK}{UTF8}{gkai}足\end{CJK}", "\begin{CJK}{UTF8}{gkai}口\end{CJK}", "\begin{CJK}{UTF8}{gkai}色\end{CJK}" are placed side by side in the middle, reorganize them into complete semantic phrases and output them on separate lines (i.e., \begin{CJK}{UTF8}{gkai}不失足于人\end{CJK}, \begin{CJK}{UTF8}{gkai}不失口于人\end{CJK}, \begin{CJK}{UTF8}{gkai}不失色于人\end{CJK}).\\
(3) Reading order and line breaks:\\
\quad(a) For regular square seals: Line break by column (right column first, then left column);\\
\quad(b) For shared characters/cross-shaped seals: Please line break and output according to the reorganized semantic phrases.\\
IV. Output Format\\
(1) Only output the recognition results. Do not include any redundant dialogue, greetings, or explanatory text. Use the "\textbackslash n" symbol for line breaks, and strictly wrap the final plain text inside the <res> and </res> tags;\\
(2) Output example: <res>text content...\begin{CJK}{UTF8}{gkai}时\end{CJK}=text content\textbackslash ntext content</res>. For example: <res>\begin{CJK}{UTF8}{gkai}晓儿之时=而分=\end{CJK}\textbackslash n\begin{CJK}{UTF8}{gkai}而不可再来\end{CJK}</res>
\end{tcolorbox}
\end{figure*}

\begin{figure*}[t]
\centering
\begin{tcolorbox}[colback=gray!5!white, colframe=gray!75!black, title=\textbf{Stele Prompt (Chinese)}, arc=2mm]
\ttfamily
\begin{CJK}{UTF8}{gkai}
这是一张墓志、碑帖或石经拓片的图片，请帮我识别图片中的文字\\
\textbf{[此处插入基础规则 Prompt]}\\
一、生僻字，异体字\\
（1）严格字形保留。对于图片中清晰可辨的字形，只要其 Unicode 编码存在，必须严格按原形输出；\\
（2）异体字替换。若字形Unicode编码不存在，但存在可打出的标准异体字，则用该异体字输出；\\
（3）若以上条件均不满足，使用<unrecognizable>标签占位。\\
二、墓志碑帖版式与“分行合书（共用字）”重组规范\\
（1）分行合书・借笔共用字重组：识别拓片内大字居中、小字分列两侧的布局结构，若两侧文字共用中心主字，严格遵循先右后左语义逻辑，将中心大字分别与左右侧文字组合，重组为完整连贯词组正常输出；\\
\quad（a）典型示例1： 若中间是大字“张”，其右上为“显”、右下为“公讳万全”，其左上为“显”、左下为“妣王氏夫人”。输出：显张公讳万全显张妣王氏夫人；\\
\quad（b）典型示例2： 若中间是大字“师”，其右上为“恩”、右下为“讳墨林”，其左上为“德”、左下为“讳广安”。输出：恩师讳墨林德师讳广安；\\
（2）双行夹注：正文旁配套小字为人物名讳、字号、履历等补充注释内容，按照从右至左顺序拼接整合，整体嵌入全角括号（）内，自然嵌入对应正文位置；\\
（3）边界区分：严格区分共用合书结构与正文行间双行夹注两种版式，不可混淆混用处理规则；\\
（4）行内空格：如果同一列的正文或批注中，上下文字之间存在明显的物理留白（如避讳抬头或段落间隔），请在对应的文字之间插入一个空格符号" "，自然嵌入到它所属的正文位置中。\\
三、输出格式\\
（1）只输出识别结果，不要包含任何多余的对话、问候或解释性文字。换行采用"\textbackslash n"符号，请将最终的纯文本严格包裹在 <res> 和 </res> 标签内部；\\
（2）输出示例：<res>文本内容（双行夹注）文本内容\textbackslash n文本内容</res>，例如：<res>大唐故李府君墓志铭\textbackslash n君讳文（字景行）世居陇西\textbackslash n曾祖讳达 隋任□州刺史\textbackslash n祖讳俨 皇朝任许州司马\textbackslash n孝子袁 考袁公子明 妣袁母崔儒 哀哀永慕\textbackslash n以开元廿载十月十日终于私第</res>
\end{CJK}
\end{tcolorbox}
\end{figure*}

\begin{figure*}[t]
\centering
\begin{tcolorbox}[colback=gray!5!white, colframe=gray!75!black, title=\textbf{Stele Prompt (English)}, arc=2mm]
\ttfamily
This is an image of an epitaph, stele rubbing, or stone classic rubbing. Please help me recognize the text in the image.\\
\textbf{[Insert Base Rule Prompt here]}\\
I. Rare and Variant Characters\\
(1) Strict character shape retention. For clearly legible character shapes in the image, as long as their Unicode encoding exists, they must be output strictly in their original form;\\
(2) Variant character replacement. If the character's Unicode encoding does not exist, but a standard typable variant character exists, output that variant character;\\
(3) If none of the above conditions are met, use the <unrecognizable> tag as a placeholder.\\
II. Stele Layout and "Split-line Combined Writing (Shared Character)" Reorganization Specifications\\
(1) Split-line combined writing / Borrowed stroke shared character reorganization: Recognize the layout structure in the rubbing where a large character is centered and small characters are distributed on both sides. If the text on both sides shares the central main character, strictly follow the right-to-left semantic logic, combine the central large character with the text on the right and left sides respectively, and reorganize them into complete and coherent phrases for normal output;\\
\quad(a) Typical example 1: If the middle is the large character "\begin{CJK}{UTF8}{gkai}张\end{CJK}", its top right is "\begin{CJK}{UTF8}{gkai}显\end{CJK}", bottom right is "\begin{CJK}{UTF8}{gkai}公讳万全\end{CJK}", top left is "\begin{CJK}{UTF8}{gkai}显\end{CJK}", and bottom left is "\begin{CJK}{UTF8}{gkai}妣王氏夫人\end{CJK}". Output: \begin{CJK}{UTF8}{gkai}显张公讳万全显张妣王氏夫人\end{CJK};\\
\quad(b) Typical example 2: If the middle is the large character "\begin{CJK}{UTF8}{gkai}师\end{CJK}", its top right is "\begin{CJK}{UTF8}{gkai}恩\end{CJK}", bottom right is "\begin{CJK}{UTF8}{gkai}讳墨林\end{CJK}", top left is "\begin{CJK}{UTF8}{gkai}德\end{CJK}", and bottom left is "\begin{CJK}{UTF8}{gkai}讳广安\end{CJK}". Output: \begin{CJK}{UTF8}{gkai}恩师讳墨林德师讳广安\end{CJK};\\
(2) Double-line interlinear notes: Small characters accompanying the main text as supplementary annotations such as names, courtesy names, and resumes should be spliced and integrated from right to left, enclosed entirely in full-width parentheses \begin{CJK}{UTF8}{gkai}（）\end{CJK}, and naturally embedded into the corresponding position in the main text;\\
(3) Boundary distinction: Strictly distinguish between the shared combined writing structure and the double-line interlinear notes layout within the main text. Do not confuse or mix the processing rules;\\
(4) Inline spaces: If there is obvious physical blank space between characters above and below in the same column of the main text or annotations (such as spacing for naming taboo or paragraph breaks), please insert a space symbol " " between the corresponding characters, naturally embedding it into its belonging main text position.\\
III. Output Format\\
(1) Only output the recognition results. Do not include any redundant dialogue, greetings, or explanatory text. Use the "\textbackslash n" symbol for line breaks, and strictly wrap the final plain text inside the <res> and </res> tags;\\
(2) Output example: <res>text content (double-line interlinear note) text content\textbackslash ntext content</res>, for example: <res>\begin{CJK}{UTF8}{gkai}大唐故李府君墓志铭\end{CJK}\textbackslash n\begin{CJK}{UTF8}{gkai}君讳文（字景行）世居陇西\end{CJK}\textbackslash n\begin{CJK}{UTF8}{gkai}曾祖讳达 隋任□州刺史\end{CJK}\textbackslash n\begin{CJK}{UTF8}{gkai}祖讳俨 皇朝任许州司马\end{CJK}\textbackslash n\begin{CJK}{UTF8}{gkai}孝子袁 考袁公子明 妣袁母崔儒 哀哀永慕\end{CJK}\textbackslash n\begin{CJK}{UTF8}{gkai}以开元廿载十月十日终于私第\end{CJK}</res>
\end{tcolorbox}
\end{figure*}

\begin{figure*}[t]
\centering
\begin{tcolorbox}[colback=gray!5!white, colframe=gray!75!black, title=\textbf{Cliff Prompt (Chinese)}, arc=2mm]
\ttfamily
\begin{CJK}{UTF8}{gkai}
这是一张摩崖石刻（刻在天然石壁上的文字）的图片，请帮我识别图片中的文字\\
\textbf{[此处插入基础规则 Prompt]}\\
一、输出格式\\
（1）只输出识别结果，不要包含任何多余的对话、问候或解释性文字。换行采用"\textbackslash n"符号，请将最终的纯文本严格包裹在 <res> 和 </res> 标签内部；\\
（2）输出示例：<res>石门颂\textbackslash n故司隶校尉犍为武阳杨君颂\textbackslash n从史位下下 某某书</res>
\end{CJK}
\end{tcolorbox}
\end{figure*}

\begin{figure*}[t]
\centering
\begin{tcolorbox}[colback=gray!5!white, colframe=gray!75!black, title=\textbf{Cliff Prompt (English)}, arc=2mm]
\ttfamily
This is an image of a cliff inscription (text carved on a natural rock face). Please help me recognize the text in the image.\\
\textbf{[Insert Base Rule Prompt here]}\\
I. Output Format\\
(1) Only output the recognition results. Do not include any redundant dialogue, greetings, or explanatory text. Use the "\textbackslash n" symbol for line breaks, and strictly wrap the final plain text inside the <res> and </res> tags;\\
(2) Output example: <res>\begin{CJK}{UTF8}{gkai}石门颂\end{CJK}\textbackslash n\begin{CJK}{UTF8}{gkai}故司隶校尉犍为武阳杨君颂\end{CJK}\textbackslash n\begin{CJK}{UTF8}{gkai}从史位下下 某某书\end{CJK}</res>
\end{tcolorbox}
\end{figure*}

\begin{figure*}[t]
\centering
\begin{tcolorbox}[colback=gray!5!white, colframe=gray!75!black, title=\textbf{Edition Prompt (Chinese)}, arc=2mm]
\ttfamily
\begin{CJK}{UTF8}{gkai}
这是一张刻本写本抄本的古籍图片，请帮我识别图片中的文字\\
\textbf{[此处插入基础规则 Prompt]}\\
一、刻本写本抄本版面元素识别与分类处理\\
（1）正文区：仅提取版框范围内主体正文与行间双行内容夹注；\\
（2）非转录区：版心/中缝（含页码、书名、鱼尾等）、版框边缘书耳（检索用篇名），放在<ignore></ignore>标签中（其中，先放版心内容，再放书耳内容，注意按列换行）；\\
（3）批注提取区：框外留白区（天头、地脚、左右侧）的手写批注、眉批，放在<note></note>标签中（其中，按逆时针方式：天头内容→左侧边注→地脚→右侧边注的顺序存放，注意按列换行）。\\
（4）双行夹注：正文行间嵌套的双行并排小字。在物理上表现为两小列字迹，两列的宽度与单行正文持平。遵循先读右列再读左列的阅读顺序，拼接后的文字放在全角括号"（）"内，自然嵌入到它所属的正文位置中；\\
（5）行内空格：如果同一列的正文或批注中，上下文字之间存在明显的物理留白（如古籍中的避讳抬头或段落间隔），请在对应的文字之间插入一个空格符号" "，自然嵌入到它所属的正文位置中。\\
二、输出格式\\
（1）只输出识别结果，不要包含任何多余的对话、问候或解释性文字。换行采用"\textbackslash n"符号，请将最终的纯文本严格包裹在 <res> 和 </res> 标签内部；\\
（2）输出示例：<res>正文内容\textbackslash n...\textbackslash n...正文上半部分\textbackslash n<ignore>版心内容1\textbackslash n版心内容2\textbackslash n...\textbackslash n书耳内容1\textbackslash n书耳内容2...</ignore>\textbackslash n正文下半部分\textbackslash n...\textbackslash n正文内容（先右再左的双列夹注内容）...\textbackslash n<note>天头内容1\textbackslash n天头内容2\textbackslash n...\textbackslash n左侧边注1\textbackslash n左侧边注2\textbackslash n...\textbackslash n地脚内容1\textbackslash n地脚内容2\textbackslash n...右侧边注1\textbackslash n右侧边注2\textbackslash n...</note></res>。\\
注意：如果图片中不存在非转录区或批注区，请直接不输出 <ignore> 或 <note> 标签。
\end{CJK}
\end{tcolorbox}
\end{figure*}

\begin{figure*}[t]
\centering
\begin{tcolorbox}[colback=gray!5!white, colframe=gray!75!black, title=\textbf{Edition Prompt (English)}, arc=2mm]
\ttfamily
This is an image of an ancient block-printed book, manuscript, or copied book. Please help me recognize the text in the image.\\
\textbf{[Insert Base Rule Prompt here]}\\
I. Layout Element Recognition and Classification Processing for Block-printed, Manuscript, and Copied Books\\
(1) Main text area: Only extract the main body text and double-line interlinear notes within the frame;\\
(2) Non-transcription area: Block center/middle seam (including page numbers, book titles, fish tails, etc.) and book ears at the edge of the frame (chapter names for retrieval) should be placed in the <ignore></ignore> tags (among them, place the block center content first, then the book ear content, paying attention to line breaks by column);\\
(3) Annotation extraction area: Handwritten annotations and top notes in the blank areas outside the frame (top margin, bottom margin, left and right sides) should be placed in the <note></note> tags (stored in a counterclockwise manner: top margin content -> left marginal notes -> bottom margin -> right marginal notes, paying attention to line breaks by column).\\
(4) Double-line interlinear notes: Double-line side-by-side small characters nested between lines of the main text. Physically, they appear as two small columns of handwriting, and the width of the two columns is equal to a single line of main text. Follow the reading order of reading the right column first and then the left column. The spliced text is placed in full-width parentheses "\begin{CJK}{UTF8}{gkai}（）\end{CJK}" and naturally embedded into its corresponding position in the main text;\\
(5) Inline spaces: If there is obvious physical blank space between characters above and below in the same column of the main text or annotations (such as spacing for naming taboo or paragraph breaks in ancient books), please insert a space symbol " " between the corresponding characters, naturally embedding it into its belonging main text position.\\
II. Output Format\\
(1) Only output the recognition results. Do not include any redundant dialogue, greetings, or explanatory text. Use the "\textbackslash n" symbol for line breaks, and strictly wrap the final plain text inside the <res> and </res> tags;\\
(2) Output example: <res>main text content\textbackslash n...\textbackslash n...upper part of main text\textbackslash n<ignore>block center content 1\textbackslash nblock center content 2\textbackslash n...\textbackslash nbook ear content 1\textbackslash nbook ear content 2...</ignore>\textbackslash nlower part of main text\textbackslash n...\textbackslash nmain text content (double-column interlinear note content, right then left)...\textbackslash n<note>top margin content 1\textbackslash ntop margin content 2\textbackslash n...\textbackslash nleft marginal note 1\textbackslash nleft marginal note 2\textbackslash n...\textbackslash nbottom margin content 1\textbackslash nbottom margin content 2\textbackslash n...right marginal note 1\textbackslash nright marginal note 2\textbackslash n...</note></res>.\\
Note: If there are no non-transcription areas or annotation areas in the image, please do not output the <ignore> or <note> tags directly.
\end{tcolorbox}
\end{figure*}

\begin{figure*}[t]
\centering
\begin{tcolorbox}[colback=gray!5!white, colframe=gray!75!black, title=\textbf{Calligraphy Prompt (Chinese)}, arc=2mm]
\ttfamily
\begin{CJK}{UTF8}{gkai}
这是一张书法、手稿或书画作品的图片，请帮我识别图片中的文字\\
\textbf{[此处插入基础规则 Prompt]}\\
一、特有符号与字形处理规范\\
（1）特殊符号（原样保留）：重文符号 "="、交换符号 "\textasciitilde"、删除三点 "、" 等古文字形符号，必须原样保留，不解读或改写；\\
（2）多书体识别：兼容篆、隶、楷、行、草五体书，同时包含甲骨文、金文（大篆）等古老书体；甲骨文字形依照权威甲骨释读规范隶定，金文字形依照商周金文权威释读规范隶定，输出对应的现代隶定文字。\\
二、特有干扰处理规范\\
（1）印章处理：图片中红色鉴藏印、作者印一律直接忽略，不识别印章内文字；仅当印章文字作为作品正文的一部分（如印文书法作品）时，再识别输出；\\
（2）墨迹与污渍区分：自动忽略纸张霉点、水渍、透字、装裱痕迹，仅提取清晰的正面主体墨迹文字；\\
（3）行内空格：同一列中存在作者刻意留白（如分段、抬格），请在对应位置插入一个空格符号 " "。\\
三、输出格式\\
（1）只输出识别结果，不要包含任何多余的对话、问候或解释性文字。换行采用"\textbackslash n"符号，请将最终的纯文本严格包裹在 <res> 和 </res> 标签内部；\\
（2）输出示例：<res>兰亭集序\textbackslash n永和九年 岁在癸丑\textbackslash n暮春之初会于会稽\textbackslash n山阴之兰亭</res>。
\end{CJK}
\end{tcolorbox}
\end{figure*}

\begin{figure*}[t]
\centering
\begin{tcolorbox}[colback=gray!5!white, colframe=gray!75!black, title=\textbf{Calligraphy Prompt (English)}, arc=2mm]
\ttfamily
This is an image of calligraphy, a manuscript, or a painting. Please help me recognize the text in the image.\\
\textbf{[Insert Base Rule Prompt here]}\\
I. Specific Symbol and Glyph Processing Standards\\
(1) Special symbols (retain as is): Ancient textual symbols such as the repetition mark "=", the exchange mark "\textasciitilde", and the three-dot deletion mark "\begin{CJK}{UTF8}{gkai}、\end{CJK}" must be retained exactly as they are, without interpretation or rewriting;\\
(2) Multi-script recognition: Compatible with the five scripts of Seal, Clerical, Regular, Running, and Cursive, while also including ancient scripts such as Oracle Bone Script and Bronze Script (Large Seal); Oracle Bone Script glyphs are transcribed according to authoritative Oracle Bone decipherment standards, and Bronze Script glyphs are transcribed according to authoritative Shang and Zhou Bronze Script decipherment standards, outputting the corresponding modern transcribed text.\\
II. Specific Interference Processing Standards\\
(1) Seal processing: Red collector's seals and author's seals in the image are strictly ignored, and the text within the seals is not recognized; only when the seal text is part of the main text of the work (such as seal script calligraphy works) should it be recognized and output;\\
(2) Distinguishing ink marks from stains: Automatically ignore paper mildew spots, water stains, bleed-through text, and mounting marks, and only extract clear main ink text on the front side;\\
(3) Inline spaces: If there is intentional blank space left by the author in the same column (such as paragraphing or elevation/taige), please insert a space symbol " " at the corresponding position.\\
III. Output Format\\
(1) Only output the recognition results. Do not include any redundant dialogue, greetings, or explanatory text. Use the "\textbackslash n" symbol for line breaks, and strictly wrap the final plain text inside the <res> and </res> tags;\\
(2) Output example: <res>\begin{CJK}{UTF8}{gkai}兰亭集序\end{CJK}\textbackslash n\begin{CJK}{UTF8}{gkai}永和九年 岁在癸丑\end{CJK}\textbackslash n\begin{CJK}{UTF8}{gkai}暮春之初会于会稽\end{CJK}\textbackslash n\begin{CJK}{UTF8}{gkai}山阴之兰亭\end{CJK}</res>.
\end{tcolorbox}
\end{figure*}

\end{document}